\documentclass[lettersize,journal]{IEEEtran}
\usepackage{amsmath,amsfonts}
\usepackage{algorithmic}
\usepackage{algorithm}
\usepackage{array}
\usepackage[caption=false,font=normalsize,labelfont=sf,textfont=sf]{subfig}
\usepackage{textcomp}
\usepackage{stfloats}
\usepackage{verbatim}
\usepackage{graphicx}
\usepackage{cite}
\usepackage{textcomp}
\usepackage{stfloats}
\usepackage{url}
\usepackage{verbatim}
\usepackage{graphicx}
\usepackage{cite}
\usepackage{bm}
\usepackage{mathrsfs}
\usepackage{multirow}
\usepackage[normalem]{ulem}
\usepackage{csquotes}
\useunder{\uline}{\ul}{}
\usepackage[colorlinks,
            linkcolor=blue,      
            anchorcolor=blue,  
            citecolor=blue,       
            ]{hyperref}

\begin{document}

\title{GaGSL: Global-augmented Graph Structure Learning via Graph Information Bottleneck}

\author{Shuangjie Li, Jiangqing Song, Baoming Zhang, Gaoli Ruan, Junyuan Xie, Chongjun Wang

\thanks{Shuangjie Li, Jiangqing Song, Baoming Zhang, Gaoli Ruan, Junyuan Xie and Chongjun Wang are with the School of Computer Science and Technology, National Key Laboratory for Novel Software Technology, Nanjing University, Nanjing 210046, China (e-mail: shuangjieli@smail.nju.edu.cn; sjq@smail.nju.edu.cn; zhangbm@smail.nju.edu.cn; glruan@smail.nju.edu.cn; jyxie@nju.edu.cn; chjwang@nju.edu.cn).}
}

\IEEEpubid{0000--0000/00\$00.00~\copyright~2021 IEEE}

\maketitle

\begin{abstract}
Graph neural networks (GNNs) are prominent for their effectiveness in processing graph data for semi-supervised node classification tasks. Most works of GNNs assume that the observed structure accurately represents the underlying node relationships. However, the graph structure is inevitably noisy or incomplete in reality, which can degrade the quality of graph representations. Therefore, it is imperative to learn a clean graph structure that balances performance and robustness. In this paper, we propose a novel method named \textit{Global-augmented Graph Structure Learning} (GaGSL), guided by the Graph Information Bottleneck (GIB) principle. The key idea behind GaGSL is to learn a compact and informative graph structure for node classification tasks. Specifically, to mitigate the bias caused by relying solely on the original structure, we first obtain augmented features and augmented structure through global feature augmentation and global structure augmentation. We then input the augmented features and augmented structure into a structure estimator with different parameters for optimization and re-definition of the graph structure, respectively. The redefined structures are combined to form the final graph structure. Finally, we employ GIB based on mutual information to guide the optimization of the graph structure to obtain the minimum sufficient graph structure. Comprehensive evaluations across a range of datasets reveal the outstanding performance and robustness of GaGSL compared with the state-of-the-art methods.
\end{abstract}


\section{Introduction}
\IEEEPARstart{G}{raph} data is pervasive in a variety of real-world scenarios, including power networks \cite{holmgren2006using}, social media \cite{krishnamurthy2008few, zhang2019your}, and computer graphics \cite{taubin1996optimal}. In these scenarios, each node with attributes represents an entity, while each edge represents the relationship between the entity pairs.
For example, in social networks, each node represents a user or individual, and they may have attributes such as personal information, interests, and professions. The edge that connects different nodes represents a friendship or following relationship, and it can be either directed or undirected.
In recent years, graph neural networks (GNNs) have emerged as a powerful approach for working with graph data \cite{velivckovic2017graph, wu2020comprehensive, wu2022graph, gilmer2017neural}, and have been widely adopted for diverse network analysis tasks, including node classification \cite{wang2020gcn, kipf2016semi}, link prediction \cite{zhang2022graph, zhang2017weisfeiler}, and graph classification \cite{zeng2021contrastive, xu2018powerful}.

Most existing GNNs rely on one basic assumption that the observed structure precisely represents the underlying node relationships.
However, this assumption does not always hold in practice, as there are multiple factors that can lead to noisy graph structure:
(1) Presence of noise and bias. Data can be collected and annotated from multiple sources. In the process of data collection and annotation, noisy connections and bias may be introduced by subjective human judgment or limitations in device precision. In some special cases (e.g., graph-enhanced applications \cite{li2020survey} and visual navigation \cite{gao2021room}), the data may lack inherent graph structure and require additional graph construction (e.g., \(k\)NN) for representation learning.
(2) Adversarial attacks on graph structure. 
The majority of current methods for adversarial attacks on graph data, including poisoning attacks \cite{demontis2019adversarial} on graph structure, concentrate on altering the graph structure, especially adding/deleting/rewiring edges \cite{xu2020adversarial}, and the original structure can be severely damaged. 
For instance, in credit card fraud detection, a fraudster might generate numerous transactions involving multiple high-credit users to conceal their identity and avoid detection by GNNs.
\IEEEpubidadjcol

\begin{figure*}[!t]
\centering
\begin{minipage}[b]{0.32\linewidth}
  \centerline{\includegraphics[width=\textwidth]{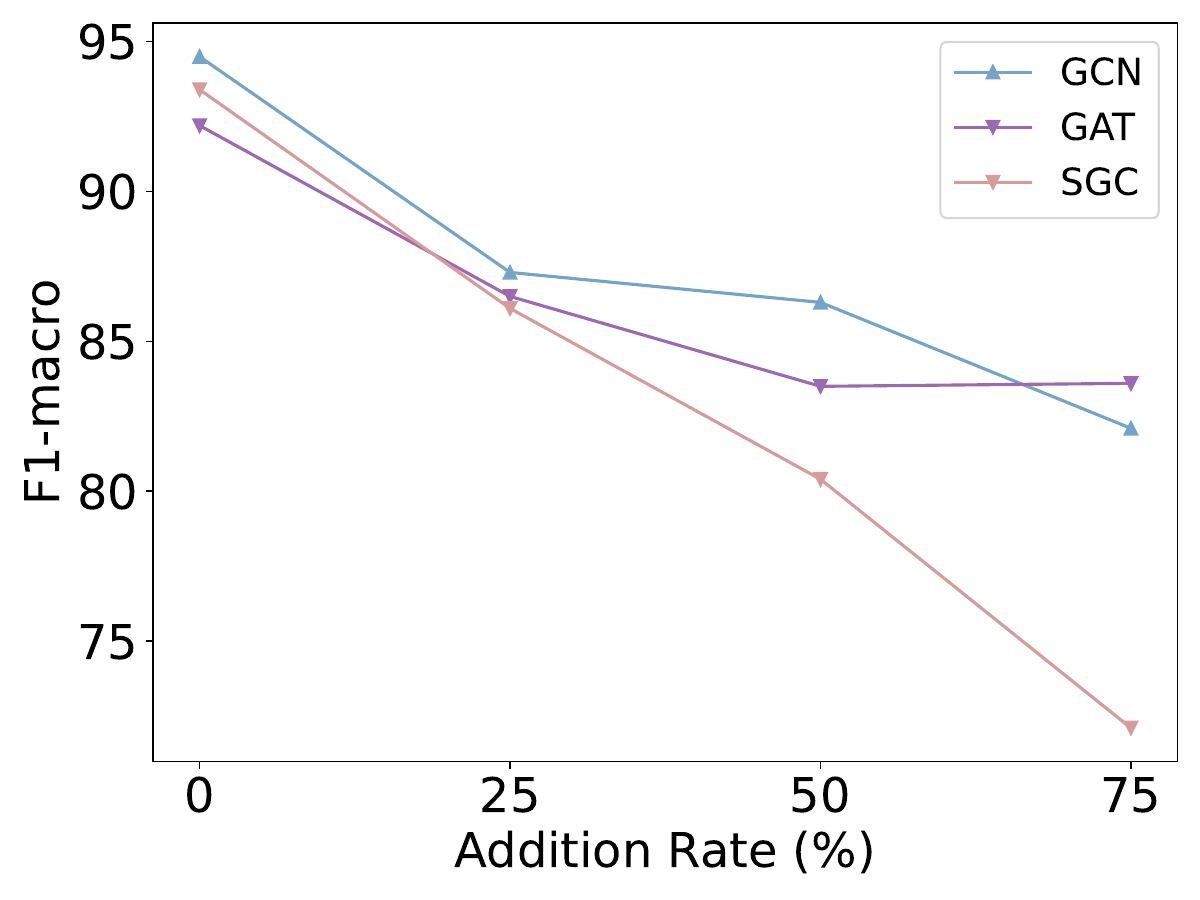}}
\end{minipage}
\hfill
\begin{minipage}[b]{0.32\linewidth}
  \centerline{\includegraphics[width=\textwidth]{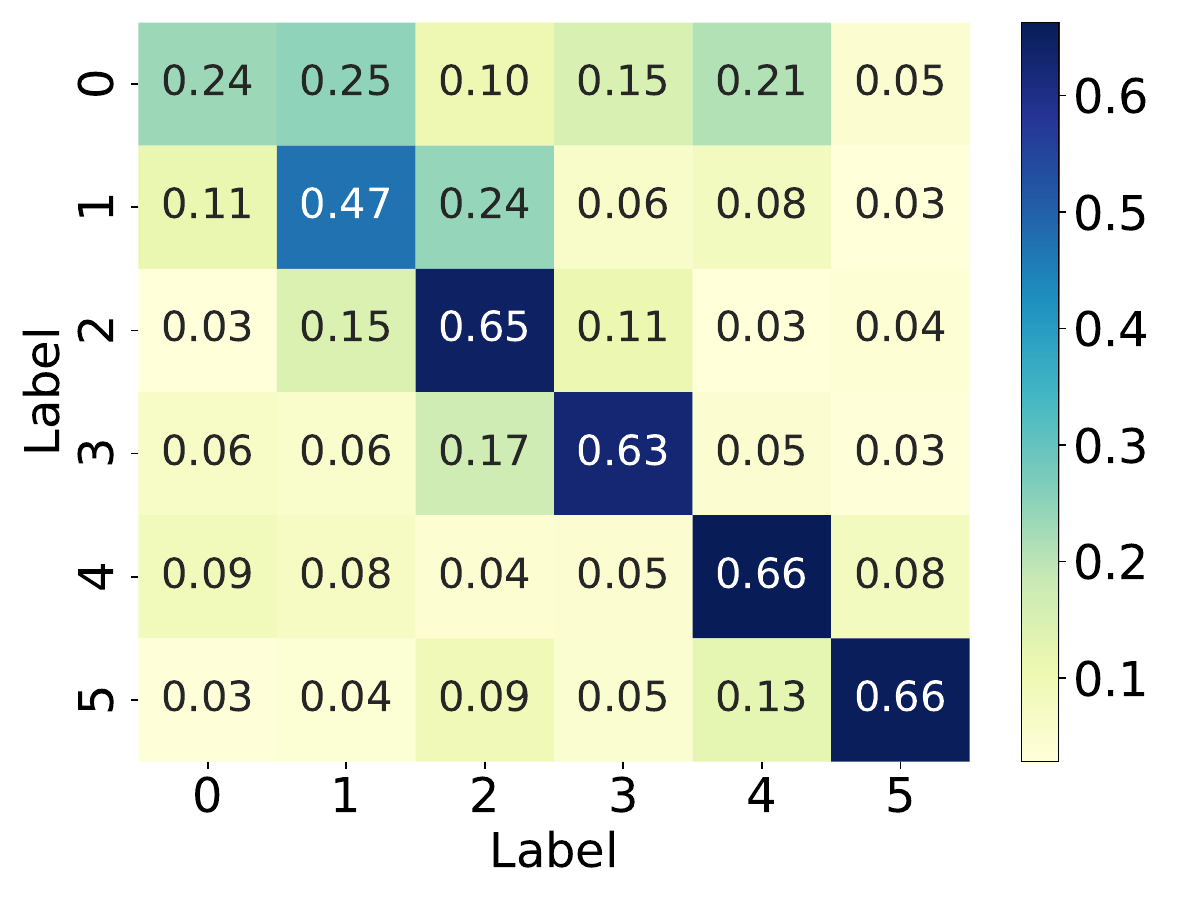}}
\end{minipage}
\hfill
\begin{minipage}[b]{0.32\linewidth}
  \centerline{\includegraphics[width=\textwidth]{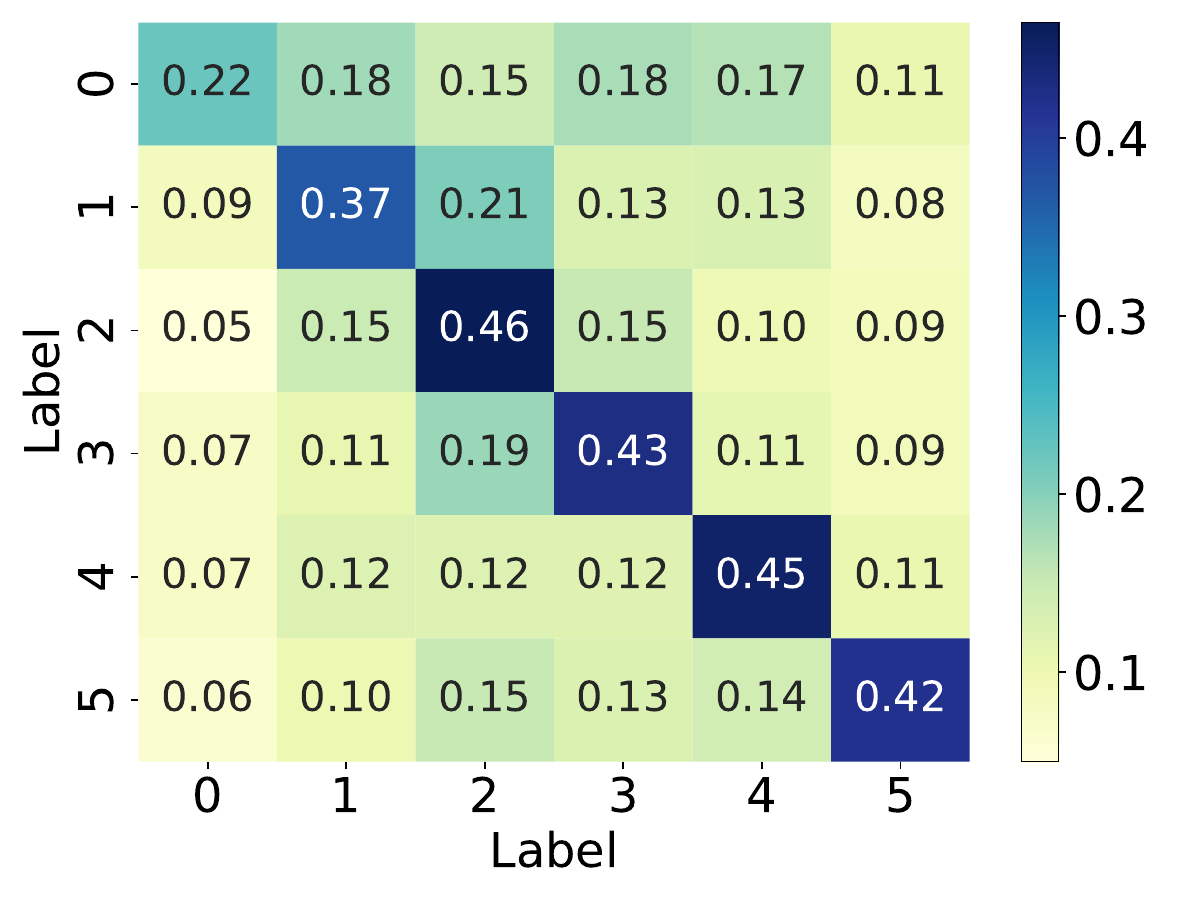}}
\end{minipage}
\caption{The performance of the models in node classification with different rates of edge addition (left), and heat maps of the probability matrices with edges added with rates of 0 (middle) and 75\% (right).}
\label{fig1}
\end{figure*}

\textit{How does the model's performance vary when the graph structure faces different levels of noise? And what is the impact on the graph structure when it is subjected to noise}?
To investigate the answer to this question, we simulated the noisy graph structure by introducing artificial edges into the graph at different proportions (i.e., 25\%, 50\%, and 75\%) of the original number of edges (simulated noise), using the polblogs \cite{jin2020graph} dataset. Additionally, we calculate the probability matrices between communities for the original graph structure, as well as the graph structure with 75\% additional edges, and draw them as heat maps.
As illustrated in Fig. \ref{fig1}, the performance of GNNs models drops dramatically with the increase of edge addition rate, with SGC \cite{wu2019simplifying} exhibiting the most significant decline.
This observation suggests that randomly adding edges has a detrimental effect on node classification performance. 
Further comparison of the middle and right plots reveals that randomly adding edges can result in connections between nodes from different communities. This inter-community connectivity decreases the distinguishing ability of nodes after GNN aggregates neighbor information, thereby leading to a decline in model performance.
In summary, the effectiveness of GNNs is heavily dependent on the underlying graph structure, while real-world graphs often suffer from missing, meaningless, or even spurious edges. This structural noise may hinder message passing and limit the generalization ability of GNNs. Therefore, there is an urgent need to explore the optimal graph structure suitable for GNNs.

In recent years, various graph structure learning (GSL) methods \cite{zhu2021deep} have emerged to handle the aforementioned problems, ensuring the performance and robustness of the models.
Nevertheless, developing effective techniques to learn an optimal graph structure for GNNs represents a technically challenging task.
(1)\textit{ How can multifaceted information be introduced to provide a more comprehensive perspective}? Relying on a single graph structure is inadequate to fully capture the complexity and diversity inherent in a graph \cite{wang2021graph}. Different perspectives of graph structure can shed light on various aspects and features within the graph. Therefore, it becomes imperative to integrate multi-perspective graph structure to acquire a more comprehensive and diverse graph structure.
Most current methods for graph structure obtain the optimal graph structure from the single original structure \cite{franceschi2019learning, jiang2019semi, jin2020graph}.
There are also methods that obtain optimal graph structure based on multiple fundamental views \cite{wang2020gcn, chen2020iterative, pei2020geom, wang2021graph}. 
As an illustrative example, Chen \textit{et al.} \cite{chen2020iterative} constructed graph structure by incorporating both the normalized adjacency matrix and the node embedding similarity matrix.
Nevertheless, it only considers feature similarity and fails to capture structural role information.
(2)\textit{ How to learn a clean graph structure for node classification tasks}?
In information theory, two key principles are Graph Information Bottleneck (GIB) \cite{wu2020graph, sun2022graph, yu2020graph} and Principle of Relevant Information (PRI) \cite{principe2010information}.
GIB and PRI represent distinct approaches to redundancy reduction and information preservation. Specifically, GIB offers an essential guideline for GSL: an optimal graph structure should contain the minimum sufficient information required for the downstream prediction task.
For example, Sun \textit{et al.} \cite{sun2022graph} advanced the GIB principle in graph classification by jointly optimizing the graph structure and graph representation.
PRI views the issue of reducing redundancy and retaining information as a balancing act. This balance is struck between reducing the entropy of the representation and its relative entropy to the original data.
For instance, a structure containing the most relevant yet least redundant information, quantified using von Neumann entropy and Quantum Jensen-Shannon divergence, was developed by Sun \textit{et al.} \cite{sun2023self}. 
However, exploring the learning of the optimal graph structure that strike a balance between performance and robustness in node classification tasks, based on information theory principle, remains an ongoing challenge.

To address the aforementioned issues, in this paper, we propose a novel \textbf{G}lobal-\textbf{a}ugmented \textbf{G}raph \textbf{S}tructure \textbf{L}earning (GaGSL) method to enhance the node classification performance and robustness based on the principle of GIB. 
GaGSL consists of global feature and structure augmentation, structure redefinition and GIB guidance.
In global feature and structure augmentation, two different techniques used to obtain augmented features and augmented structure, respectively. Details are presented in subsection \ref{augment}.
In structure redefinition, we introduce a structure estimator to appropriately refine the graph structure. This involves reallocating weights to the graph adjacency matrix elements based on node similarity. 
Then, the redefined structures are integrated to form the final structure. More information is provided in subsection \ref{redefine}.
In GIB guidance, we aim to maximize the mutual information (MI) between node labels and node embeddings $\bm Z^*$ based on the final graph structure, while simultaneously imposing constraints on the MI between $\bm Z^*$ and the node embeddings $\bm Z_{r1}$ or $\bm Z_{r2}$ based on the redefined structures. To effectively evaluate the MI, we employ an MI calculator based on the InfoNCE loss \cite{oord2018representation}. More elaboration is given in subsection \ref{GIB_guidence}.
Finally, we employ a cyclic optimization scheme to iteratively update the model parameters. Details are presented in subsection \ref{iter_optim}. Our contributions can be summarized as follows:
\begin{itemize}
  \item We propose a Global-augmented GSL method, GaGSL, which is guided by GIB and aims at obtaining the most compact structure.
  This endeavor seeks to strike a finer balance between performance and robustness.
  \item To alleviate the limitations of relying solely on a single graph structure, we integrate augmented features and augmented structure to obtain a more global and diverse graph structure.
  \item The evaluation of our proposed GaGSL method is conducted on eight benchmark datasets, and the experimental findings convincingly showcase the effectiveness of GaGSL. Notably, GaGSL exhibits superior performance when contrasted to state-of-the-art GSL methods, and this advantage is particularly pronounced when the method is applied to datasets that have been subjected to attacks.
\end{itemize}

The rest of the paper is organized as follows. In section \ref{Rw} we briefly introduce the related works.
Before presenting our research methodology in Section \ref{Me}, we give the relevant notations and backgrounds in Section \ref{Nb}.
In Section \ref{Exp}, we present and discuss the results of our experiments on benchmark datasets. 
Finally, Section \ref{Con} outlines the conclusions drawn from this work and discusses potential avenues for future research.

\section{\label{Rw}Related works}
\noindent 
Aligned with the focus of our study, we provide a concise review of the two research areas most pertinent to our work - Graph Neural Networks and graph structure learning.

\subsection{Graph Neural Networks}
\noindent GNNs have emerged as a prominent approach due to their effectiveness in working with graph data. These GNN models can be broadly categorized into two main groups: spectral-based methods and spatial-based methods.

Spectral-based methods aim to identify graph patterns in the frequency domain, leveraging the sound mathematical precepts of Graph Signal Processing (GSP) \cite{hammond2011wavelets, qiao2022rpt}. Bruna \textit{et al.} \cite{bruna2013spectral} first extended the convolution to general graphs using a Fourier basis, treating the filter as a set of learnable parameters and considering graph signals with multiple channels. but eigendecomposition requires \(O(n^3)\) computational complexity. In order to reduce the computational
complexity, Defferrard \textit{et al.} \cite{defferrard2016convolutional} and Kipf \textit{et al.} \cite{kipf2016semi} made several approximations and simplifications.
Defferrard \textit{et al.} \cite{defferrard2016convolutional} defined fast localized convolutional filters on graphs based on Chebyshev polynomials. 
Kipf \textit{et al.} \cite{kipf2016semi} further simplified ChebNet via a localized first-order approximation of spectral graph convolutions. 
Despite being spectral-based, GCN can also be viewed through a spatial perspective. In this context, GCN operates by aggregating feature information from the local neighborhood of each node. Recent studies have progressively improved upon GCN \cite{kipf2016semi} by exploring alternative symmetric matrices.
For example,
The adaptive Graph Convolution Network (AGCN) \cite{li2018adaptive} proposed a Spectral Graph Convolution layer with graph Laplacian Learning (SGC-LL), which efficiently adapts the graph topology according to the data and learning task context. Dual Graph Convolutional Network (DGCN) \cite{zhuang2018dual} hamilton2017inductivedesigned a dual neural network structure to encode both local and global consistency. Other spectral graph convolutions have also been proposed.
Graph Wavelet Neural Network (GWNN) \cite{xu2019graph} defines the convolution operator via wavelet transform, which avoids matrix eigendecomposition and provides good interpretability with its local and sparse graph wavelets. 
Simple Spectral Graph Convolution (S\(^2\)GC)\cite{zhu2021simple} derives a variant of GCN based on a modified Markov Diffusion Kernel. It achieves a balance between low- and high-pass filter bands to capture both global and local contexts of each node.

Conversely, spatial-based methods draw inspiration from the message passing mechanism employed in Recurrent Graph Neural Network (RecGNN) \cite{gallicchio2010graph, dai2018learning}. They directly define graph convolution in the spatial domain as transforming and aggregating local information.
The Neural Network for Graphs (NN4G) \cite{micheli2009neural} is the first work towards spatial-based convolutional graph neural networks (ConvGNNs). Unlike RecGNNs, 
NN4G utilizes a combinatorial neural network architecture where each layer has independent parameters to learn the mutual dependencies of the graph.
This method allows the extension of a node's neighborhood through the progressive construction of the architecture.
To identify a particularly effective variant of the general approach and apply it to the task of chemical property prediction, the message passing neural network (MPNN) \cite{gilmer2017neural} provides a unified model for spatial-based ConvGNNs.
In this framework, graph convolutions are treated as a message-passing process, where information is directly transmitted between nodes along the edges.
The graph isomorphism network (GIN) \cite{xu2018powerful} identifies that previous methods based on MPNNs lack the ability to differentiate between distinct graph structures based on the embeddings they generate. 
To address this limitation, GIN proposed a theoretical framework for analyzing the expressive capabilities of GNNs to capture different graph structures. 
Obtaining the full size of a node's neighborhood is inefficient since the number of neighbors a node has can span a wide range, from as few as one to possibly more than a thousand or more.
Hamilton \textit{et al.} \cite{hamilton2017inductive} generated representations by sampling and aggregating features from a node’s local neighborhood.
Velivckovic \textit{et al.} \cite{velivckovic2017graph} assumed that the contributions of neighboring nodes to the central node are neither identical as in GraphSage nor predetermined as in GCN.
Velivckovic \textit{et al.} \cite{velivckovic2017graph} assigned distinct edge weights based on node features during aggregation process. 
GCN is primarily inspired by the latest deep learning methods and therefore may inherit unnecessary complexity and redundant computation.
Wu \textit{et al.} \cite{wu2019simplifying} reduced complexity and computation by successively removing nonlinearities and collapsing weight matrices between consecutive layers.

Many other graph neural network models can be reviewed in recent surveys \cite{wu2020comprehensive, zhou2020graph}. But almost all these GNNs assume that the observed structure accurately reflects the underlying node relationships, which considerably constrains their ability to manage uncertainty in the graph topology.

\subsection{Graph Structure Learning}
\noindent GSL attempts to approximate a better structure for the original graph, which is not a newly emerged topic and has roots in prior works in network science \cite{lusher2013exponential, martin2016structural}. GSL methods can be generally classified into three categories: metric learning approaches, probabilistic modeling approaches and direct optimization approaches.

Metric learning methods polish the graph structure by learning a metric function that evaluates the similarity between pairs of node representations. For example, Zhang \textit{et al.} \cite{zhang2020gnnguard} and Wang \textit{et al.} \cite{wang2020gcn} leveraged cosine similarity to model the edge weights.
Zhang \textit{et al.} \cite{zhang2020gnnguard} detected fake edges with different features and labels between nodes and mitigates their negative impact on prediction by removing these edges or reducing their weight in neural messaging.
Wang \textit{et al.} \cite{wang2020gcn} employed a distinct method to derive the final node embeddings. It diffuses node features across the original graph and integrates the representations from both the generated feature graph and the original input graph using an attention mechanism.
Note that Chen \textit{et al.} \cite{chen2020iterative} constructed the graph via a multi-head self-attention network, which is an end-to-end graph learning framework for iteratively optimizing graph structure and node embeddings. 
The core idea of IDGL \cite{chen2020iterative} is that better node embedding is beneficial to capture better graph structure.

Probabilistic modeling approaches assume that graph is generated through a sampling process from certain distributions, and they use learnable parameters to model the probability of sampling edges.
Franceschi \textit{et al.} \cite{franceschi2019learning} modeled the edges between each pair of nodes by sampling from Bernoulli distributions with learnable parameters, and presented GSL as a two-layer programming problem, which was the first work in probabilistic modeling.
The approach proposed by Zhang \textit{et al.} \cite{zhang2019bayesian} employed Monte Carlo dropout to sample the learnable model parameters multiple times for each generated graph.
Meanwhile, the method developed by Wang \textit{et al.} \cite{wang2021graph}  processed multi-view information, such as multi-order neighborhood similarity, as observations of the optimal graph structure, and then derived the final graph structure based on Bayesian inference.

Direct optimization approaches consider the graph adjacency matrix as trainable parameters, which are tuned jointly alongside the primary GNN parameters.
Yang \textit{et al.} \cite{yang2019topology} aimed to leverage a given class label to both refine the graph structure and optimize the parameters of the GNN simultaneously.
Jin \textit{et al.} \cite{jin2020graph} focused on investigating key graph properties like sparsity, low rank, and feature smoothness, with the goal of designing more robust graph neural network models.

GSL methods can also be roughly split into single view and multiple fundamental views methods. 
However, these methods are insufficient to explore the theoretical guidance of learning optimal structure.
A more comprehensive review of GSL can be found in a recent study \cite{zhu2021deep}.

\section{\label{Nb}Notations and Backgrounds}
\noindent In this section, we present the notations and backgrounds related to this paper. 

\subsection{Notions}
\noindent Let \(G=(\bm{A}, \bm{X})\) be a graph with adjacency matrix \(\bm{A}\in \mathbb{R}^{|\bm{V}|\times|\bm{V}|}\) and node feature matrix \(\bm{X}\in \mathbb{R}^{|\bm{V}|\times F}\), where \(\bm{V}:=\{v_i\}_{i=1}^N\) denotes node set.
Following the commonly adopted semi-supervised node classification setting, only a small portion of nodes, denoted as \(\bm{V}_L:=\{v_i\}_{i=1}^M\), have associated labels available, represented as \(\bm{Y}:=\{y_i\}_{i=1}^M\), where \(y_i\) corresponds to the label of node \(v_i\).
\(\bm{L} = \bm{I}_N-\bm{D}^{-1/2}\bm{A}\bm{D}^{-1/2}\) denotes the normalized graph Laplacian matrix, where \(\bm{I}_N\) represents the identity matrix, and \(\bm{D} \in \mathbb{R}^{|\bm{V}|\times|\bm{V}|}\) is a diagnoal degree matrix with \(\bm{D}_{i,i}=\sum_j\bm{A}_{i,j}\).

Given an input graph \(G=(\bm{A}, \bm{X})\) and partial node labels \(\bm{Y}\), the objective of GSL for GNNs is to jointly learn an optimal graph structure and the GNN model parameters, with the aim of enhancing the node classification performance for unlabeled nodes.
The main notations used in this paper are summarized in Table \ref{table1}.
\subsection{Graph Neural Network}
\noindent
Modern GNNs stack multiple graph convolution layers to learn high-level node representations. The convolution operation usually consists of two steps: aggregation and update, which are respectively represented as follows:
\begin{equation}
\bm{m}_i^{(l)} = \mathrm{AGGREGATE}^{(l)}(\bm{h}_j^{(l)}, {v_j}\in \mathcal{N}({v_i})) 
\end{equation}
\begin{equation}
\bm{h}_i^{(l+1)} = \mathrm{UPDATE}^{(l)}(\bm{h}_i^{(l)}, \bm{m}_i^{(l)}) 
\end{equation}
where \(\bm{m}_i^{(l)}\) and \(\bm{h}_i^{(l)}\) are the message vector and the hidden embedding of node \(v_i\) at the \textit{l}-th layer, respectively.
The set \(\mathcal{N}({v_i})\) consists of nodes that are adjacent to node \(v_i\).
If \(l=0\), then \(\bm{h}_i^{(0)} = \bm{x}_i\). \(\mathrm{AGGREGATE}^{(l)}(\cdot)\) and \(\mathrm{UPDATE}^{(l)}(\cdot)\) are characterized by the specific model, respectively. In an \textit{L}-layer network, the final embedding \(\bm{h}_i^{(L)}\) is fed to a linear fully connected layer for the classification task.

\subsection{Graph Information Bottleneck}
\noindent Inspired by the Information Bottleneck (IB), Wu \textit{et al.} \cite{wu2020graph} proposed GIB principle to optimize node-level representations \(\bm Z\) to to capture the minimal sufficient information within the input graph data \(G = (\bm A, \bm X)\) required for predicting the target \(\bm Y\).
The objective is defined as the following optimization:
\begin{equation}
   {\rm arg}\mathop{\min}_{\bm Z} -I(\bm Z; \bm Y )+\beta I(\bm Z; G)
\end{equation}

Intuitively, the first term \(-I(\bm Z; \bm Y )\) encourages the representations \(\bm Z\) to be maximally informative about the target (sufficient).
The second term \((\bm Z; G)\) serves to prevent \(\bm Z\) from obtaining extraneous information from the data that is not pertinent to predicting the target (minimal).
The Lagrangian multiplier \(\beta\) trading off sufficiency and minimality. Existing works based on GIB primarily focus on graph-level tasks \cite{yu2020graph, yu2022improving, sun2022graph}. In this paper, we specifically discuss GIB for node classification tasks.

\begin{table}[]
    \centering
    \caption{SUMMARY OF THE MAIN NOTATIONS IN THIS PAPER}
    \resizebox{\linewidth}{!}{
    \begin{tabular}{l||l}
    \hline
    Notions & Descriptions               \\ \hline
    \hline
    \(\bm{A}\)       & original structure                     \\ \hline
    \(\bm{A}^*\)     & the optimized original    \\ \hline
    \(\bm{X}\)       & original features                     \\ \hline
    \(\bm{x}_i\)       & feature vector of node \(v_i\)                \\ \hline
    \(\bm{V}\)      &   node set                         \\ \hline
    \(\bm{D}\)        &   diagnoal degree matrix                        \\ \hline
    \(\bm{D_{ii}}\)        &   The \(i\)th diagonal element of the degree matrix \(\bm{D}\)                      \\ \hline
    \(G^*\)       &   the optimized graph                         \\ \hline
     \(G_{r1}\)        &   the redefined graph 1                         \\ \hline
    \(G_{r2}\)        &    the redefined graph 2                        \\ \hline
    \(\bm{v}_i\)  & the embedding of node \(v_i\) in graph  \(G_{r1}\)  \\ \hline
    \(\bm{u}_i\)  & the embedding of node \(v_i\) in graph  \(G^*\)  \\ \hline
    \(s(\bm{u}_i,\bm{v}_i)\)  & cosine similarity of \(\bm{u}_i\) and \(\bm{v}_i\) \\ \hline
    \(g_s\)  & filter kerner \\ \hline
    \(F\)       & the dimension of feature   \\ \hline
    \(N\)       & the number of nodes \\ \hline
    \(\beta\)  &  Lagrangian multiplier \\
    \hline
    \end{tabular}
    }
\label{table1}
\end{table}
\section{\label{Me}Methodology}
\noindent In this section, we illustrate the proposed streamlined GSL model GaGSL guided by GIB. We begin with the overview of GaGSL in subsection \ref{overview}. Subsequently, we will detail the 3 main parts of GaGSL (i.e., global feature and structure augmentation in subsection \ref{augment}, structure redefinition in subsection \ref{redefine}, and GIB guidance in subsection \ref{GIB_guidence}). 
Finally, we detail the process of joint iterative optimization in subsection \ref{iter_optim}.

\begin{figure*}[!t]
    \centering
    \includegraphics[width=0.85\linewidth]{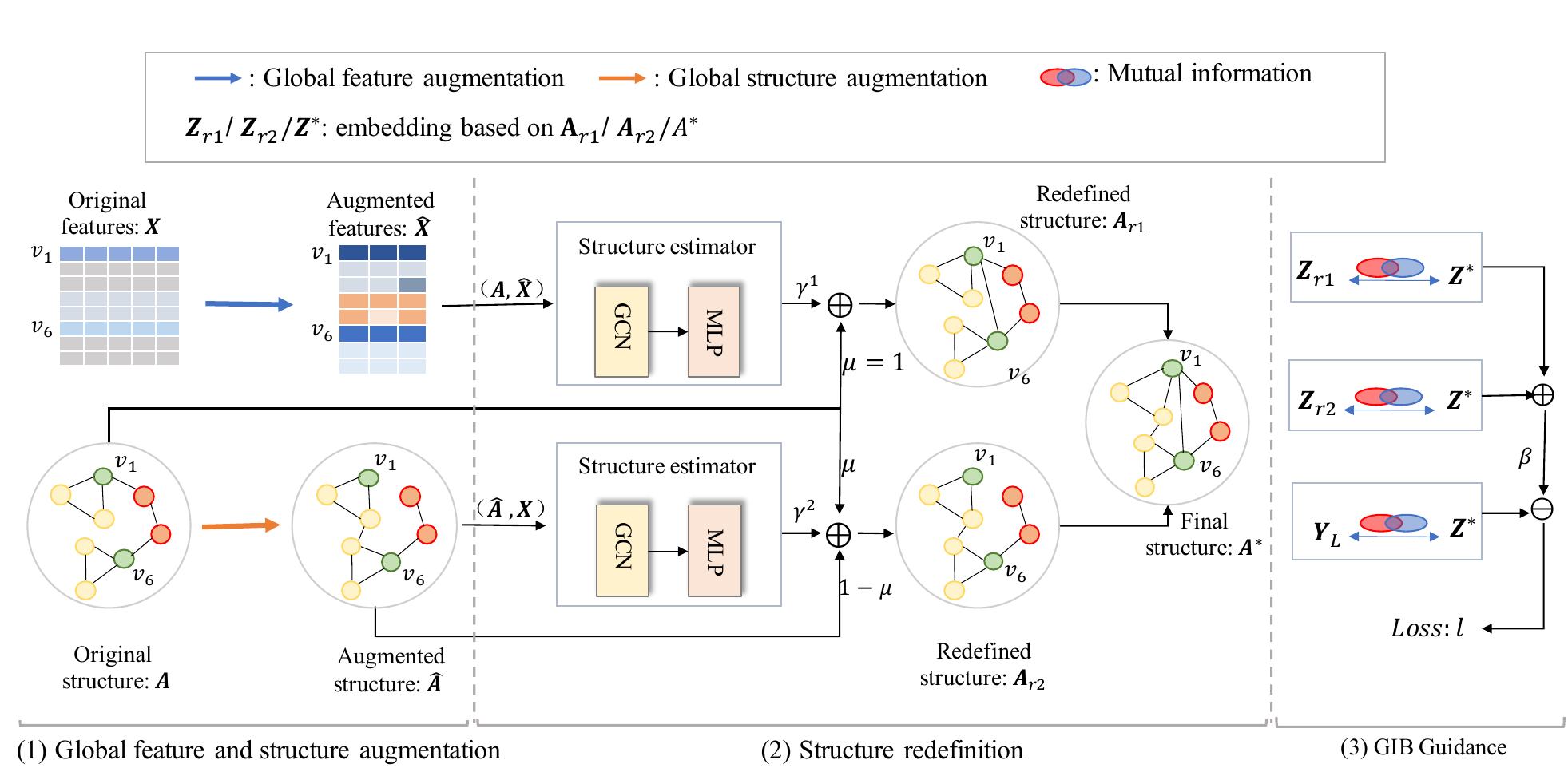}
    \caption{Overview of GaGSL. Given original features \(\bm X\) and structure \(\bm A\) as input, GaGSL consists of the following three parts: (1) Global feature and structure augmentation: augmented features \(\hat{\bm X}\) and augmented structure \(\hat{\bm A}\) are obtained by global feature augmentation and global structure augmentation, respectively; (2) Structure redefinition: based on the augmented graph data \((\bm A, \hat{\bm X})\) or \((\hat{\bm A},\bm X)\), the graph structure is redefined using a structure estimator; (3) GIB Guidance: learning minimal sufficient graph structure guided by GIB.}
    \label{fig2}
\end{figure*}
\subsection{\label{overview}Overview}

\noindent Most existing GNNs assume that the observed graph structure accurately reflects the true relationships between nodes. However, this assumption often fails in real-world scenarios where graphs are typically noisy.
This pitfall in the observed graph can lead to a rapid decline in GNN performance.
Thus, a natural and promising approach is to jointly optimize the graph structure and the GNN model parameters in an integrated manner.
In this paper, the proposed GaGSL aims to learn compact and informative graph structure for node classification tasks guided by the principle of GIB. 
Fig. \ref{fig2} provides an overview of the GaGSL model.

\subsection{\label{augment}Global Feature and Structure Augmentation}
\noindent Most of the current research in GSL is predominantly conducted from a single structure. However, this single-structure approach is susceptible to biases that can result in an incomplete understanding of the entire graph structure, ultimately limiting the performance and robustness of the model. 
To mitigate this concern, 
we employ a two-pronged approach that combines global feature augmentation and global structure augmentation. This aims to comprehensively understand the graph structure from multiple perspectives.

\subsubsection{Global Feature Augmentation}
Nodes located in different regions of a graph may exhibit similar structural roles within their local network topology, and recognizing these roles is essential for understanding network organization. 
For example, nodes \(v_1\) and \(v_6\) in Fig. \ref{fig2} exhibit similar structural roles despite being far apart in the graph.
Following previous works \cite{sun2023self,donnat2018learning}, we use spectral graph wavelets and empirical characteristic function to generate structural embedding for every node.

The filter kernel \(g_s\) is characterized by a scaling parameter \(s\) that controls the reach of the diffusion process, where greater \(s\) promotes wider-ranging diffusion.

In this paper, we employ the heat kernel \(g_{s}(\lambda)=e^{-\lambda s}\). The spectral graph wavelet \(\Psi(v_i)\) centered around node \(v_i\) is given by an \(N\)-dimensional vector:
\begin{equation}
    \Psi_s(v_i) = \bm{U} {\rm Diag}(g_{s}(\lambda_1),g_{s}(\lambda_2),...,g_{s}(\lambda_N))\bm{U}^T\bm{\delta}_i
\end{equation}
where \(\lambda_i\) and \(\bm U\) denote the eigenvalue and the eigenvector of the graph Laplacian \(\bm L\), respectively. \(\bm{\delta}_i\) is the one-hot vector of node \(v_i\).

To address the node mapping problem, we regard the wavelets as probability distributions and describe them using empirical characteristic functions.
Following \cite{sun2022graph}, the empirical characteristic function of \(v_i\) is:
\begin{equation}
\label{eq6}
    \Phi_s(v_i, t)=\frac{1}{N}\sum_{n=1}^N e^{-i\Psi_s(v_i)t}
\end{equation}

Lastly, the structural embedding \(\bm h_s(v_i)\) of node \(v_i\) is obtained by sampling a 2-dimensional parametric function (as defined in Eq. (\ref{eq6})) at \(d\) different points \(\{t_1,t_2,...,t_d\}\), and then concatenating the resulting values:

\begin{equation}
    \bm{h}_s(v_i)=[{\rm Re}(\Phi_s(v_i, t_i)),{\rm Im}(\Phi_s(v_i, t_i))]_{t_1,t_2,...,t_d}
\end{equation}

To effectively encode both the local and global structural roles of a node,
we consider a set of different scales \(\{s_1,s_2,...,s_m\}\) to integrate the information from different neighborhood radii to obtain a multi-scale structural role embedding:
\begin{equation}
\label{eq_hvi}
    \bm{h}({v_i})={\rm Concat}(\bm h_{s_1}(v_i),\bm h_{s_2}(v_i),...,\bm h_{s_m}(v_i))
\end{equation}

We can construct an augmented features matrix \(\hat{\bm X}\) by Eq. (\ref{eq_hvi}).

\subsubsection{Global Structure Augmentation}
In addition to structural role embeddings, we also augment structure from another perspective. Specifically, we employ the widely-used diffusion matrix to capture the global relationships between nodes.
We utilize Personalized PageRank (PPR) \cite{gasteiger2018predict}, which provides a comprehensive representation of global structure.
The PPR has a closed-form solution given as follows:
\begin{equation}
\label{eqPpr}
    \hat{\bm{A}}=\alpha(\bm{I}_N-(1-\alpha)\bm{D}^{-1/2}\bm{A}\bm{D}^{-1/2})^{-1}
\end{equation}
where, \(\alpha\) denotes the restart probability. Note that \(\bm{S}\) could be dense, thus we just keep 5 edges for each node on some datasets corresponding to the top 5 most affinity nodes on some datasets.

\subsection{\label{redefine}Structure Redefinition}
\noindent Given two graph \(G_{af} = (A, \hat{\bm{X}})\) and \(G_{as} = (\hat{\bm{A}},\bm X)\), to further capture the complex associations and semantic similarities between nodes, we perform a structure estimator for each graph. Specifically, for graph \(G_{af}\), we conduct a layer of GCN \cite{kipf2016semi} followed by a MLP layer:
\begin{equation}
\label{eq9}
    \bm{H}^1={\rm GCN}(\bm{A}, \hat{\bm{X}})
\end{equation}
\begin{equation}
\label{eq10}
    w_{ij}^1={\rm MLP}([\bm{h}_i^1||\bm{h}_j^1])
\end{equation}
where \(w_{ij}^1\) denotes the weight between node \(v_i\) and node \(v_j\), \(\bm{h}_i^1\) and \(\bm{h}_j^1\) are the embeddings of node \(v_i\) and node \(v_j\), respectively.
Here, to mitigate resource consumption in terms of both space and time, we only estimate the \(h\)-order neighbors of each node. Then, \(w_{ij}^1\) are normalized via sotfmax function to get the final weight:
\begin{equation}
\label{eq11}
    \bm{S}_{ij}^1=\frac{exp(w_{ij}^1)}{\sum exp(w_{ik}^1)}
\end{equation}

We can construct a similarity matrix \(\bm{S}^1\) by Eq. (\ref{eq11}). 
The original graph structure \(\bm{A}\) carries relatively rich information. 
Ideally, the learned graph structure \(\bm{S}^1\) can complement the original graph structure \(\bm{A}\), creating an optimized graph for GNNs that enhances performance on the downstream task \cite{chen2020iterative}.
Thus, the matrix \(\bm{A}\) is combined with the similarity matrix \(\bm{S}^1\) to get the redefined structure \(\bm{A}_{r1}\):
\begin{equation}
\label{eq12}
    \bm{A}_{r1}= \bm{A} + \gamma^1*\bm{S}^1
\end{equation}
where \(\gamma^1 \in [0,1]\) is combination coefficient. 
Similarly, we can obtain the redefined structure \(\bm{A}_{r2}\) for graph \(G_{af}\):
\begin{equation}
\label{eq13}
    \bm{A}_{r2}=\mu * \bm{A} + (1-\mu)*\hat{\bm{A}} +\gamma^2*\bm{S}^2
\end{equation}
where \(\gamma^2\), \(\mu \in [0,1]\) are combination coefficients. Note that when using Eq. (\ref{eq11}) based on the diffusion matrix, top-\(k\) neighbors are selected for each node based on the PPR values.

After employing structure redefinition, we can obtain two corresponding graphs \(G_{r1} = (\bm{A}_{r1}, \bm X)\) and \(G_{r2} = (\bm{A}_{r2}, \bm X)\).

\subsection{\label{GIB_guidence}GIB Guidance}
\noindent In this section, the question we would like to answer is how to get the optimal structure \(\bm A^*\) for node classification tasks? And how to guide the training of \(\bm A^*\) so that it is the minimal sufficient? In order to avoid introducing new parameters and to simplify the model as much as possible, we obtain the final structure by applying the average function, with the inputs being two redefined structures:
\begin{equation}
    \bm{A}^*=\frac{1}{2}(\bm{A}_{r1}+\bm{A}_{r2})
    \label{eq14}
\end{equation}

Please review that we aims to learn minimum sufficient graph structure for node classification tasks. In other words, we want the learned representations \(\bm Z^*\) based on \(G^*=(\bm{A}^*, \bm X)\) to contain only labeled information, while filtering out label-irrelevant noise. To the end, we use GIB to guide the training of \(\bm{A}^*\) so that it is the minimal sufficient.
We presume that no information is lost during this process, following the standard practice of MI estimation \cite{tian2020makes}.
Therefore, we have \(I(G^*;G_{r1}) \approx I(\bm Z^*;G_{r1})\) and \(I(G^*;\bm Y) \approx I(\bm Z^*;\bm Y)\). For the sake of convenience in the following, we replace \(I(\bm Z^*,\bm Y)\) with \(I(G^*;\bm Y)\) and \(I(\bm Z^*;G_{r1})\) with \(I(G^*;G_{r1})\).
The objectives of GaGSL are as follows:
\begin{equation}
\label{eq15}
    {\rm arg}\mathop{\min}_{G^*}-I(G^*;\bm{Y})+\beta^1 I(G^*;G_{r1})
\end{equation}
\begin{equation}
\label{eq16}
    {\rm arg}\mathop{\min}_{G^*}-I(G^*;\bm{Y})+\beta^2 I(G^*;G_{r2})
\end{equation}

By adding Eq. (\ref{eq15}) and Eq. (\ref{eq16}), we obtain:
\begin{equation}
\label{eq17}
     {\rm arg}\mathop{\min}_{G^*}-I(G^*;\bm{Y})+\beta^1 I(G^*;G_{r1})+\beta^2 I(G^*;G_{r2})
\end{equation}

We can simplify Eq. (\ref{eq17}) by letting \(\beta^1=\beta^2=\beta\):
\begin{equation}
\label{eq18}
    {\rm arg}\mathop{\min}_{G^*}-I(G^*;\bm{Y})+\beta(I(G^*;G_{r1})+I(G^*;G_{r2}))
\end{equation}
where the first term \(-I(G^*;\bm{Y})\) is used to encourage \(G^*\) to contain maximum information about the labels \(\bm{Y}\). The second term \(I(G^*;G_{r1})+I(G^*;G_{r2})\)
encourages \(G^*\) to contain as little irrelevant information from \(G_{r1}\) and \(G_{r2}\) as possible, which is label-irrelevant for predicting the target.

The non-Euclidean nature of graph data makes it challenging to estimate the MI in Eq. (\ref{eq18}) accurately \cite{paninski2003estimation}.
Therefore, we introduce a variational upper bound of \(-I(G^*;\bm{Y})\), and use the InfoNCE \cite{chen2020simple,he2020momentum} approximation to calculate \(I(G^*;G_{r1})\) and \(I(G^*;G_{r2})\). First, we examine the prediction term \(-I(G^*;\bm{Y})\).

\textbf{Proposition 4.1} (Upper bound of \(-I(G^*;\bm{Y})\)). Given graph \(G\) with label \(\bm{Y}\) and \(G^*\) learned from \(G\), we have
\begin{equation}
\label{eq19}
    -I(G^*;\bm{Y})\leq-\iint p(\bm{Y},G^*){\rm log}(q_\theta(\bm{Y}|G^*))d\bm{Y} dG^* + H(\bm{Y})
\end{equation}
where \(q_\theta(\bm{Y}|G^*)\) is the variational approximation of the \(p(\bm{Y}|G^*)\). We have the following proof based on Sun \textit{et al.} \cite{sun2022graph}:
\begin{equation}
\label{eq20}
  \begin{aligned}
    -I(G^*;\bm{Y}) & =-\iint p(\bm{Y},G^*) \log \frac{p(\bm{Y}, G^*)}{p(\bm{Y}) p(G^*)} d\bm{Y} dG^* \\
    & =-\iint p(\bm{Y}, G^*) \log \frac{p(\bm{Y}|G^*)}{p(\bm{Y})} d\bm{Y} dG^*
  \end{aligned}
\end{equation}
Since \(p(\bm{Y}|G^*)\) is intractable, let \(q_\theta(\bm{Y}|G^*)\) is the variational approximation of the true posterior \(p(\bm{Y}|G^*)\). According to the non-negativity of Kullback Leiber divergence:
\begin{equation}
\label{eq21}
    \begin{aligned}
    & D_{KL}(p(\bm{Y}|G^*)||q_{\theta}(\bm{Y}|G^*)) \ge 0 \Longrightarrow \\
    & \int p(\bm{Y}|G^*)\log P(\bm{Y}|G^*)d\bm{Y} \ge \\
    &\int p(\bm{Y}|G^*)\log q_{\theta}(\bm{Y}|G^*)d\bm{Y}
  \end{aligned}
\end{equation}

Plug Eq. (\ref{eq20}) into Eq. (\ref{eq21}), then we have
\begin{equation}
  \begin{aligned}
    -I(G^*;\bm{Y}) & \leq -\iint p(\bm{Y},G^*) \log \frac{q_{\theta}(\bm{Y}|G^*)}{p(\bm{Y})} d\bm{Y} dG^* \\
    & =-\iint p(\bm{Y}, G^*) \log{q_{\theta}(\bm{Y}|G^*)} d\bm{Y} dG^* +H(\bm{Y})
  \end{aligned}
\end{equation}
where \(H(\bm{Y})\) is the entropy of label \(\bm{Y}\), which can be ignored in optimization procedure.

It is not trivial to compute the integral directly. To optimize the objective in Eq. (\ref{eq19}), we approximate the integral by Monte Carlo sampling \cite{shapiro2003monte} of all training samples, so that we have:
\begin{equation}
  \begin{aligned}
    -I(G^*;\bm{Y})&\leq -\iint p(\bm{Y}, G^*) \log{q_{\theta}(\bm{Y}|G^*)} d\bm{Y} dG^*\\ 
    & \approx \frac{1}{N}\sum_{i=1}^N\lbrace -\log q_{\theta}(\bm{Y}_i|\bm{Z}_i^*)\rbrace:=\mathcal{L}_{cls}(\bm{Z}^*,\bm{Y})
  \end{aligned}
\end{equation}
where \(\bm{Z}_i^*\) represents the embedding of node \(v_i\) and \(q_{\theta}(\bm{Y}|G^*)\) provides the label distribution of the learned graph \(G^*\), which can be modeled as a classifier. The classification loss, \(\mathcal{L}_{cls}\), is chosen to be the cross-entropy loss.

Then, we introduce how to approximate the mutual information \(I(G^*;G_{r1})\) and \(I(G^*;G_{r2})\) using InfoNCE. The InfoNCE loss \cite{oord2018representation} has been shown to maximize a lower bound of MI. Here, we design a MI calculator. Specifically, for \(G_{r1}\) we conduct a layer of GCN \cite{kipf2016semi} followed by a shared two-layer MLP:
\begin{equation}
\label{eq24}
    \bm{H}^{r1}={\rm GCN}(\bm{A}_{r1}, \bm{X})
\end{equation}
\begin{equation}
\label{eq25}
    \bm{H}_p^{r1}={\rm MLP}(\bm{\bm{H}}^{r1})
\end{equation}
where \(\bm{H}^{r1}\) and \(\bm{H}_p^{r1}\) are the node representations obtained from GCN and MLP, respectively.
Similarly, we can get the projected embedding \(\bm{H}_p^{r2}\) and \(\bm{H}_p^{*}\) of \(G_{r2}\) and \(G^*\), respectively. For readability purposes, we represent the embedding of node \(v_i\) in graph  \(G^*\) as \(\bm{u}_i\), and the embedding of \(v_i\) in graph \(G_{r1}\) as \(\bm{v}_i\). Then, the InfoNCE loss of graph \(G_{r1}\) and graph \(G^*\) is given following the GCA \cite{zhu2021graph}:
\begin{equation}
\label{eq26}
    \mathcal{L}(G^*,G_{r1})=\frac{1}{2B}\sum_{i=1}^B[\ell(\bm{u}_i,\bm{v}_i) +\ell(\bm{v_i},\bm{u_i})]
\end{equation}
where \(B\) is the number of nodes that randomly sampled. The pairwise objective for each positive pair \((\bm{u}_i,\bm{v}_i)\) is defined as follows: 
\begin{equation}
\label{eq27}
    \ell (\bm{u}_i,\bm{v}_i)=\log \frac{e^{s(\bm{u}_i,\bm{v}_i)/\tau}}{e^{s(\bm{u}_i,\bm{v}_i)/\tau} + \sum_{k\neq i}e^{s(\bm{u}_i,\bm{v}_i)/\tau}}
\end{equation}
where \(s(\bm{u}_i,\bm{v}_i)\) is the cosine similarity and \(\tau\) is temperature coefficient. Similarly, we can calculate loss \(\mathcal{L}(G^*,G_{r2})\).
\subsection{\label{iter_optim}Iterative Optimization}
\noindent 
Optimizing the parameters \(\Theta\) of the structure estimator, \(\Phi\) of the MI calculator, and \(\Omega\) of the classifier simultaneously is challenging. The interdependence among them further complicates this process. In this study, we employ an alternating optimization approach to iteratively update \(\Theta\), \(\Phi\), and \(\Omega\) inspired by Wang \textit{et al.} \cite{wang2021graph}.

\subsubsection{Update \(\Phi\)}
The parameters involved in Eq. (\ref{eq24}) and (\ref{eq25}) are regarded as the parameters \(\Phi\) of MI calculator.
To encourage \(G^*\) to contain as little label-irrelevant information from \(G_{r1}\) and \(G_{r2}\) as possible, the objective function used to optimize the MI calculator is presented as follows:
\begin{equation}
\label{eq28}
     \mathcal{L}_{MI}=\mathcal{L}(G^*,G_{r1})+\mathcal{L}(G^*,G_{r2})
\end{equation}
\subsubsection{Update \(\Omega\)}
We can obtain the final learned structure \(\bm{A}^*\) by Eq. (\ref{eq14}). Then we employ two-layer of GCN \cite{kipf2016semi} to obtain node representations.
\begin{equation}
\label{eq29}
     \bm{Z}^*={\rm GCN}(\bm{A}^*, \bm{X})
\end{equation}

The parameters involved in Eq. (\ref{eq29}) are collectively considered as the classifier's parameters \(\Omega\), and the cross-entropy loss is utilized for optimization:

\begin{equation}
\label{eq30}
    \mathcal{L}_{cls}=\sum_{v\in V}{\rm Cross\text{-}Entropy}(\bm{Z}_i^*, y_v)
\end{equation}

\subsubsection{Update \(\Theta\)}
After training the classifier and MI calculator, 
we proceed with the continuous optimization of the structure estimator parameters \(\Theta\). Guided by GIB, the resulting loss function is as follows:
\begin{equation}
\label{eq31}
    \mathcal{L}=\mathcal{L}_{cls}-\beta\mathcal{L}_{MI}
\end{equation}
where \(\beta\) is a balance parameter trading off sufficiency
and minimality. The first term \(\mathcal{L}_{cls}\) is to motivate 
\(G^*\) to contain maximal information about the labels \(Y\) in order to improve the performance on the predicted target. The intention of the second term \(\mathcal{L}_{MI}\) is to minimize the information in \(G^*\) from \(G_{r1}\) and \(G_{r2}\) that is label-irrelevant for predicting the target.

\subsubsection{Training Algorithm}
\noindent Based on the previously described update and inference rules, the training algorithm for GaGSL is outlined in Algorithm \ref{alg1}. Specifically, the algorithm begins by initializing all the parameters of GaGSL.
In lines 4-8, GaGSL updates the parameters \(\Theta\) of structure estimator. In line 9, the redefined structures are combined. In lines 11-14, the parameter \(\Phi\) of the MI calculator is optimized. Finally, in lines 16-18, the classifier parameters \(\Omega\) are updated. Through this substitution and iterative updating, the graph structure \(\bm{A}^*\) and the better parameters promote each other.

\begin{algorithm}[htb]
\caption{Model training for GaGSL.}\label{alg:alg1}
\begin{algorithmic}[1]
\REQUIRE ~~\\ 
$ \text{adjacency matrix } \bm{A}, \text{feature matrix } \bm{X}, \text{labels } \bm{Y}_L$,\\
total epochs $T$, training classifier epochs $T_{c}$, \\
training structure estimator epochs $T_{v}$, \\
training MI calculator epochs $T_{m}$
\ENSURE ~~\\ 
learned graph structure $\bm{A^*}, \text{ GCN parameters } \Omega$

\STATE Initialize params: $\Theta, \Omega, \Phi; $ 
\STATE Initialize views: $G_{i1}, G_{i2}; $ 
\STATE $ \textbf{for } i \leftarrow \{1,2,...,T\} \textbf{ do} $ 
\STATE \hspace{0.5cm} $ \textbf{for } i \leftarrow \{1,2,...,T_v\} \textbf{ do} $ 
\STATE \hspace{1cm} $ \text{calculate } \bm{S}_{ij}^1 , \bm{S}_{ij}^2 \text{ with Eq. (\ref{eq9}), (\ref{eq10}), and (\ref{eq11})}; $
\STATE \hspace{1cm} $ \text{obtain } \bm{A}_{r1} \text{ with Eq. (\ref{eq12})}; $
\STATE \hspace{1cm} $ \text{obtain } \bm{A}_{r2} \text{ with Eq. (\ref{eq13})}; $
\STATE \hspace{1cm} $ \text{update } \Theta \text{ with Eq. (\ref{eq31})}; $
\STATE \hspace{0.5cm} $ \textbf{end}$

\STATE \hspace{0.5cm} $ \bm{A}^* = 1/2(\bm{A}_{r1}+\bm{A}_{r2});$
\STATE \hspace{0.5cm} $ \textbf{for } i \leftarrow \{1,2,...,T_m\} \textbf{ do} $ 
\STATE \hspace{1cm} $ \text{calculate } \mathcal{L}(G^*,G_{r1}) \text{ with Eq. (\ref{eq24})-(\ref{eq26})};$
\STATE \hspace{1cm} $ \text{calculate } \mathcal{L}(G^*,G_{r2}) \text{ with Eq. (\ref{eq24})-(\ref{eq26})};$
\STATE \hspace{1cm} $ \text{update } \Phi \text{ with Eq. (\ref{eq28})}; $
\STATE \hspace{0.5cm} $ \textbf{end}$
\STATE \hspace{0.5cm} $ \textbf{for } i \leftarrow \{1,2,...,T_c\} \textbf{ do} $ 
\STATE \hspace{1cm} $\text{calculate node embeddings } \bm{Z}^* \text{ with Eq. (\ref{eq29})};$
\STATE \hspace{1cm} $ \text{update } \Omega \text{ with Eq. (\ref{eq30})}; $

\STATE \hspace{0.5cm} $\textbf{end}$
\STATE $\textbf{end}$
\STATE $\textbf{return } \bm{A}^*, \Omega$
\end{algorithmic}
\label{alg1}
\end{algorithm}

\section{\label{Exp}Experiments}
\noindent 
In this section, we carry out a comprehensive evaluation to assess the effectiveness of the proposed GaGSL model. We first compare the performance of GaGSL against several state-of-the-art methods on the semi-supervised node classification task. Additionally, we perform an ablation study to verify the importance of each component within the GaGSL model.
Then, we analyze the robustness of GaGSL. Finally, we present the graph structure visualization, hyper-parameter sensitivity and values in the learned structure.
\subsection{Experiment Setup}
\subsubsection{Datasets}
The eight datasets we employ consist of four academic networks (Cora, Citeseer, Wiki-CS, and MS Academic (MS)), three non-graph datasets (Wine, Breast Cancer (Cancer), and Digits) that are readily available in scikit-learn \cite{pedregosa2011scikit}, and a blog graph dataset Polblogs. Table \ref{table2} provides a summary of the statistical information about these datasets. It is important to note that, for the non-graph datasets, we adopt the approach described in \cite{chen2020iterative} and construct a \(k\)NN graph as the original adjacency matrix.
\begin{table}[]
    \centering
    \caption{The statistics of the datasets}
    \resizebox{\linewidth}{!}{
    \begin{tabular}{llllll}
    \hline
    Dataset  & \#Nodes & \#Edges & \#Features & \#Classes & \#Train/\#Val/\#Test \\ \hline
    Wine     & 178     & 3560    & 13         & 3         & 10/20/148            \\
    Cancer   & 569     & 22760   & 30         & 2         & 10/20/539            \\
    Digits   & 1797    & 43128   & 64         & 10        & 50/100/1647          \\ \hline
    Polblogs \cite{jin2020graph} & 1222    & 33428   & 1490       & 2         & 121/123/978          \\
    Cora \cite{kipf2016semi}    & 2708    & 5429    & 1433       & 7         & 140/500/1000         \\
    Citeseer \cite{kipf2016semi} & 3327    & 9228    & 3703       & 6         & 120/500/1000         \\
    Wiki-CS \cite{mernyei2020wiki} & 11701   & 291039  & 300        & 10        & 200/500/1000         \\
    MS \cite{gasteiger2018predict}      & 18333   & 163788  & 6850       & 15        & 300/500/1000         \\ \hline
    \end{tabular}
    }
\label{table2}
\end{table}

\subsubsection{Baselines}
To demonstrate the effectiveness
of our proposed method, we compare the proposed GaGSL with two categories of baselines: three classical GNN models (GCN \cite{kipf2016semi}, GAT \cite{velivckovic2017graph}, SGC \cite{wu2019simplifying})
and four GSL based methods (Pro-GNN \cite{jin2020graph},
IDGL \cite{chen2020iterative}, GEN \cite{wang2021graph}, PRI-GSL \cite{sun2023self}). The details are given as follows.

\textit{a) GCN:} It directly encodes the graph structure using a neural network, and trains on a supervised target for all labeled nodes. This neural network employs an efficient layer-wise propagation rule, which is derived from a first-order approximation of spectral graph convolutions.

\textit{b) GAT:} It introduces an attention-based mechanism for classifying nodes in graph-structured data. By stacking layers, it enables nodes to incorporate features from their neighbors and implicitly assigns different weights to different nodes in the neighborhood. Additionally, this model can be directly applied to inductive learning problems.

\textit{c) SGC:} It alleviates the excessive complexity of GCNs by iteratively eliminating nonlinearities between GCN layers and consolidating weights into a single matrix.

\textit{d) Pro-GNN:} To defend against adversarial attacks, it iteratively eliminates adversarial structure by preserving the graph low rank, sparsity, and feature smoothness, while maintaining the intrinsic graph structure.

\textit{e) IDGL:} Building on the principle that improved node embeddings lead to better graph structure, it introduces an end-to-end graph learning framework for the joint iterative learning of graph structure and embeddings. Additionally, it frames the graph learning challenge as a similarity metric learning problem and employs adaptive graph regularization to manage the quality of the learned graph.

\textit{f) GEN:} It is a graph structure estimation neural network composed of two main components: the structure model and the observation model.
The structure model characterizes the underlying graph generation process, while the observation model incorporates multi-order neighborhood information to accurately infer the graph structure using Bayesian inference techniques.

\textit{g) PRI-GSL:} It is an information-theoretic framework for learning graph structure, grounded in the Principle of Relevant Information to manage structure quality. It incorporates a role-aware graph structure learner to develop a more effective graph that maintains the graph's self-organization.

\subsubsection{Implementation}
For three classical GNN models (GCN, GAT, SGC), we use the corresponding Pytorch Geometric library implementations \cite{fey2019fast}. 
For four GSL based methods (Pro-GNN, IDGL, GEN, and PRI-GSL), we utilize the source codes provided by the authors and adhere to the settings outlined in their original papers, with careful tuning.
For different datasets, we follow the original splits on training/validation/test. For the proposed GaGSL, we use Adam \cite{kingma2014adam} optimizer and adopt 16 hidden dimensions.
We set the learning rate for the classifier and the MI calculator to a fixed value of 0.01, while tuning it for the structure estimator across the values \{0.1, 0.01, 0.001\}.
On the MS dataset we set combination coefficient \(\mu\) to 0. On the other datasets we set \(\mu\) to 1. We test the combination coefficients \(\gamma^1\) and \(\gamma^2\) in the range \{0.1, 0.5\}. The dropout for classifier is chosen form \{0.3, 0.5, 0.7, 0.9\}, and the dropout for MI calculator is turned amongst \{0.2, 0.4, 0.6, 0.8\}. 

\begin{table*}[]
\centering
\caption{Quantitative results (\(\%\pm\sigma\)) on semi-supervised node classification task. The best-performing models are bolded and runners-up are underlined. The "-" symbol indicates that experiments could not be run due to memory problems.}
\resizebox{\linewidth}{!}{
\begin{tabular}{llllllllll}
\hline
Dataset                                   & Metric   & GCN            & GAT       & SGC            & Pro-GNN        & IDGL              & GEN               & PRI-GSL        & \textbf{GaGSL}             \\ \hline
\multicolumn{1}{c}{\multirow{3}{*}{Wine}} & AUC      & 99.4\(\pm\)0.1       & 98.1\(\pm\)1.8  & 99.4\(\pm\)0.1       & {\ul 99.6\(\pm\)0.1} & 99.6\(\pm\)1.1          & 98.6\(\pm\)0.7          & 99.1\(\pm\)0.3       & \textbf{99.8\(\pm\)0.1} \\
\multicolumn{1}{c}{}                      & F1-macro & {\ul 97.5\(\pm\)0.5} & 93.7\(\pm\)3.1  & 97.2\(\pm\)0.6       & 97.0\(\pm\)0.3       & 95.6\(\pm\)1.7          & 95.4\(\pm\)1.7          & 91.9\(\pm\)2.4       & \textbf{97.9\(\pm\)0.3} \\
\multicolumn{1}{c}{}                      & F1-micro & 97.4\(\pm\)0.5       & 93.4\(\pm\)4.2  & 97.2\(\pm\)0.7       & {\ul 97.5\(\pm\)0.3} & 95.3\(\pm\)1.8          & 95.1\(\pm\)1.9          & 91.5\(\pm\)2.5       & \textbf{97.9\(\pm\)0.3} \\ \hline
\multirow{3}{*}{Cancer}                   & AUC      & 96.7\(\pm\)0.4       & 95.8\(\pm\)1.4  & 97.4\(\pm\)0.2       & 97.8\(\pm\)0.2       & 97.9\(\pm\)0.9          & 97.8\(\pm\)0.3          & {\ul 98.4\(\pm\)0.2} & \textbf{98.8\(\pm\)0.2} \\
                                          & F1-macro & 91.5\(\pm\)0.7       & 89.3\(\pm\)2.1  & 91.3\(\pm\)0.4       & 93.3\(\pm\)0.5       & 91.9\(\pm\)2.7          & 93.5\(\pm\)0.3          & {\ul 94.2\(\pm\)0.6} & \textbf{94.6\(\pm\)0.3} \\
                                          & F1-micro & 92.0\(\pm\)0.7       & 89.8\(\pm\)2.2  & 91.7\(\pm\)0.4       & 93.8\(\pm\)0.5       & 92.5\(\pm\)2.5          & 93.9\(\pm\)0.3          & {\ul 94.6\(\pm\)0.5} & \textbf{95.0\(\pm\)0.2} \\ \hline
\multirow{3}{*}{Digits}                   & AUC      & 98.5\(\pm\)1.7       & 99.0\(\pm\)0.2  & {\ul 99.2\(\pm\)0.1} & 98.1\(\pm\)0.2       & 98.9\(\pm\)0.4          & 98.8\(\pm\)0.4          & 98.3\(\pm\)0.3       & \textbf{99.5\(\pm\)0.1} \\
                                          & F1-macro & 88.9\(\pm\)1.9       & 90.2\(\pm\)0.7  & 89.4\(\pm\)0.2       & 89.7\(\pm\)0.3       & {\ul 90.4\(\pm\)1.2}    & 92.0\(\pm\)0.5          & 90.3\(\pm\)0.8       & \textbf{92.7\(\pm\)0.4} \\
                                          & F1-micro & 89.1\(\pm\)1.8       & 90.3\(\pm\)0.7  & 89.6\(\pm\)0.2       & 89.8\(\pm\)0.3       & 90.4\(\pm\)1.2          & {\ul 92.0\(\pm\)0.5}    & 90.4\(\pm\)0.8       & \textbf{92.8\(\pm\)0.4} \\ \hline
\multirow{3}{*}{Polblogs}                 & AUC      & 98.4\(\pm\)0.0       & 97.2\(\pm\)1.3  & {\ul 98.5\(\pm\)0.0} & 98.1\(\pm\)0.2       & 98.0\(\pm\)0.5          & 98.0\(\pm\)0.5          & 98.4\(\pm\)0.1       & \textbf{98.6\(\pm\)0.1} \\
                                          & F1-macro & {\ul 95.3\(\pm\)0.3} & 92.0\(\pm\)2.6  & 94.8\(\pm\)0.0       & 94.6\(\pm\)0.7       & 94.4\(\pm\)1.6          & 95.2\(\pm\)0.8          & 95.1\(\pm\)0.4       & \textbf{95.9\(\pm\)0.2} \\
                                          & F1-micro & 95.0\(\pm\)0.3       & 92.0\(\pm\)2.7  & 94.8\(\pm\)0.1       & 94.6\(\pm\)0.7       & 94.5\(\pm\)1.5          & {\ul 95.2\(\pm\)0.8}    & 95.1\(\pm\)0.5       & \textbf{95.9\(\pm\)0.2} \\ \hline
\multirow{3}{*}{Cora}                     & AUC      & 93.7\(\pm\)0.7       & 92.5\(\pm\)0.7 & 96.1\(\pm\)0.1      & 96.9\(\pm\)0.9      & {\ul 97.0\(\pm\)0.3}    & 94.0\(\pm\)1.9          & 95.8\(\pm\)0.3       & \textbf{97.3\(\pm\)0.3} \\
                                          & F1-macro & 74.3\(\pm\)1.8       & 70.3\(\pm\)0.9  & 78.4\(\pm\)0.2      & 78.8\(\pm\)2.6      & {\ul 80.4\(\pm\)1.3}    & 80.1\(\pm\)1.3          & 75.0\(\pm\)0.4       & \textbf{82.3\(\pm\)1.1} \\
                                          & F1-micro & 75.1\(\pm\)2.3       & 71.4\(\pm\)1.1  & 79.4\(\pm\)0.2       & 79.8\(\pm\)2.6       & {\ul 82.2\(\pm\)2.6}    & 91.4\(\pm\)2.0          & 76.7\(\pm\)0.7       & \textbf{83.8\(\pm\)1.2} \\ \hline
\multirow{3}{*}{Citeseer}                 & AUC      & 87.5\(\pm\)1.2       & 89.4\(\pm\)0.2  & 89.9\(\pm\)0.2       & 88.5\(\pm\)0.3       & \textbf{91.4\(\pm\)0.4} & 90.1\(\pm\)1.7          & 88.6\(\pm\)0.2       & {\ul 90.6\(\pm\)0.5}    \\
                                          & F1-macro & 63.2\(\pm\)0.4       & 63.3\(\pm\)1.2  & 66.4\(\pm\)0.5       & 63.1\(\pm\)0.7       & \textbf{69.4\(\pm\)0.4} & 69.4\(\pm\)1.4          & 64.5\(\pm\)0.4       & {\ul 68.8\(\pm\)0.9}    \\
                                          & F1-micro & 67.1\(\pm\)0.3       & 66.2\(\pm\)1.2  & 70.6\(\pm\)0.1       & 65.6\(\pm\)0.8       & 71.9\(\pm\)0.3          & \textbf{72.7\(\pm\)1.3} & 67.6\(\pm\)0.6       & {\ul 72.1\(\pm\)1.1}    \\ \hline
\multirow{3}{*}{Wiki-CS}                  & AUC      & 91.3\(\pm\)0.5       & 91.0\(\pm\)0.6  & {\ul 94.0\(\pm\)0.0} & 93.3\(\pm\)0.3       & 91.8\(\pm\)0.2          & 92.6\(\pm\)1.2          & -              & \textbf{96.1\(\pm\)2.7} \\
                                          & F1-macro & 62.5\(\pm\)1.7       & 62.0\(\pm\)1.7  & 67.5\(\pm\)0.2       & 64.8\(\pm\)2.0       & {\ul 69.1\(\pm\)1.1}    & 68.4\(\pm\)0.3          & -              & \textbf{73.6\(\pm\)2.4} \\
                                          & F1-micro & 67.4\(\pm\)1.6       & 62.3\(\pm\)1.4  & 71.5\(\pm\)0.2       & 68.3\(\pm\)1.2       & {\ul 72.7\(\pm\)0.8}    & 71.1\(\pm\)0.9          & -              & \textbf{75.0\(\pm\)2.4} \\ \hline
\multirow{3}{*}{MS}                       & AUC      & 95.8\(\pm\)0.7       & 98.0\(\pm\)0.2  & {\ul 98.8\(\pm\)0.1} & -              & 96.5\(\pm\)0.3          & 89.0\(\pm\)0.8          & -              & \textbf{99.6\(\pm\)0.1}        \\
                                          & F1-macro & 78.7\(\pm\)3.6       & 80.7\(\pm\)1.1  & 90.0\(\pm\)0.1       & -              & 78.5\(\pm\)1.8          & \textbf{92.0\(\pm\)0.6} & -              & {\ul 90.2\(\pm\)0.5}           \\
                                          & F1-micro & 80.8\(\pm\)3.4       & 83.4\(\pm\)0.8  & 91.9\(\pm\)0.2       & -              & 82.9\(\pm\)0.9          & \textbf{98.8\(\pm\)0.3} & -              & {\ul 92.3\(\pm\)0.4}           \\ \hline
\end{tabular}}
\label{table3}
\end{table*}

\begin{table}[]
\centering
\caption{Ablation study on various datasets. w/o FA (SA/SE/GIB) denotes that the feature augmentation (structure augmentation/structure estimator/GIB guidance) component is turned off.}
\resizebox{\linewidth}{!}{
\begin{tabular}{lllll}
\hline
Methods & Cancer   & Polblogs & Citeseer & MS       \\ \hline
w/o FA  & 93.9\(\pm\)0.3        & 95.6\(\pm\)0.2        & 67.8\(\pm\)1.7        & 89.9\(\pm\)0.2        \\
w/o SA  & 93.6\(\pm\)0.4        & 94.0\(\pm\)0.6        & 65.6\(\pm\)3.0        & 89\(\pm\)0.5        \\
w/o SE  & 93.8\(\pm\)0.5        & 95.3\(\pm\)0.3        & 67.7\(\pm\)1.8        & 89.1\(\pm\)0.4        \\
w/o GIB & 94.6\(\pm\)0.6        & 95.2\(\pm\)0.1        & 67.2\(\pm\)0.8        & 90\(\pm\)0.7        \\ \hline
GaGSL   & \textbf{95.4\(\pm\)0.3} & \textbf{96.0\(\pm\)0.2} & \textbf{68.8\(\pm\)0.9} & \textbf{90.2\(\pm\)0.5} \\ \hline
\end{tabular}
}
\label{table4}
\end{table}

\subsection{Node Classification}
\noindent In this section, we assess the proposed GaGSL on semi-supervised node classification, with the results presented in Table \ref{table3}. 
To conduct a comprehensive evaluation of our model, we utilize three widely-used performance metrics: F1-macro, F1-micro, and AUC. We report the mean and standard deviation of the results obtained over 10 independent trials with varying random seeds. Based on these evaluation results, we draw the following observations:

\begin{itemize}
    \item Compared with other baselines, the proposed GaGSL can obtain better or competitive results on all datasets, which shows that our cleverly designed GSL framework can effectively improve the node classification performance.
    \item GaGSL demonstrates a significant performance improvement compared to the baseline GCN. Specifically, across all datasets, GaGSL achieves an AUC that is 0.4\%-4.8\% higher than GCN, an F1-macro score that is 0.4\%-11.5\% higher than GCN, and an F1-micro score that is 0.5\%-11.5\% higher than GCN. This observation suggests that GaGSL effectively mitigates the bias caused by a single perspective, enabling it to learn an appropriate graph structure by considering multi-perspective.
    \item Our performance improvement compared to other GSL methods demonstrates the effectiveness of utilizing GIB for guiding GSL. The learned graph structure contains more valuable information and less noise, making it better suited for node classification tasks. Although GaGSL was the runner-up on the Citeseer dataset, it has an extremely weak disadvantage over the winner, possibly due to data imbalance or the presence of labeling noise.
\end{itemize}

\subsection{Ablation Study}
\noindent Here, we describe the results of the ablation study for the different modules in the model. We report F1-macro and standard deviation results over 5 independent trials using different random seeds. 
According to the ablation study results presented in Table \ref{table4}, the model's performance exhibits a significant decline when the structure augmentation (SA) component is removed.
The substantial drop in performance observed without the SA component suggests that the structure augmentation mechanism is highly important and plays a crucial role in enabling the model to effectively capture salient features.
Notably, turning off any of the individual components results in a significant decrease in the model's performance across all evaluated datasets, which underscores the effectiveness and importance of these components. Furthermore, the results highlight the benefits of leveraging GIB to guide the model training process.

\begin{figure*}[htbp]
    \centering
    \begin{minipage}[b]{0.3\textwidth}
        \centering
        \includegraphics[width=\textwidth]{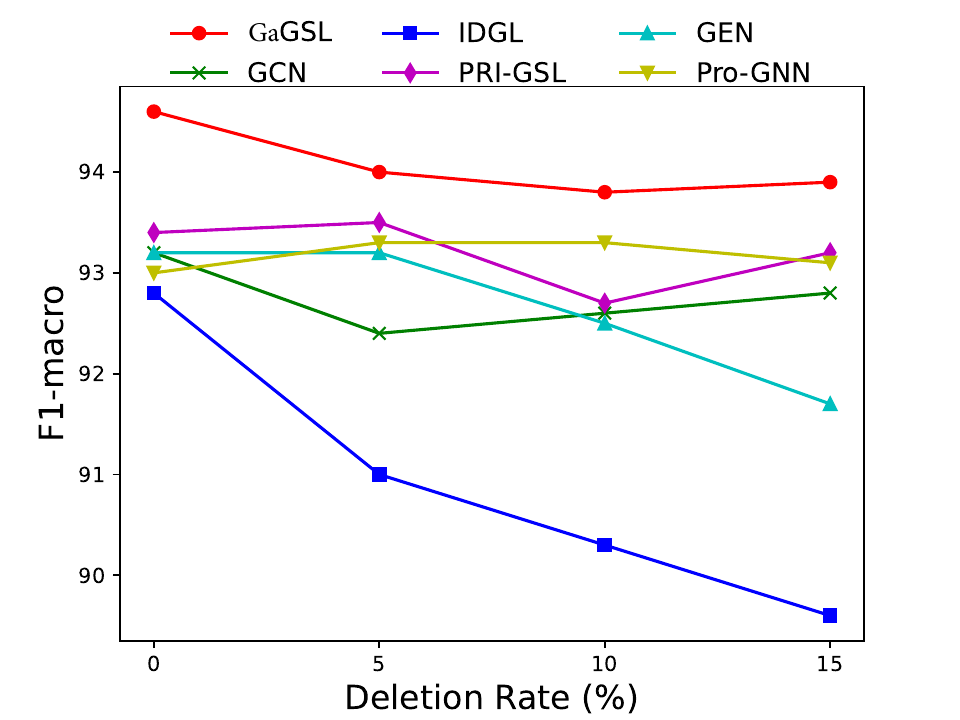}
        \centerline{\small{(a) Cancer}}
        \label{fig:dele1}
    \end{minipage}
    \begin{minipage}[b]{0.3\textwidth}
    \centering
    \includegraphics[width=\textwidth]{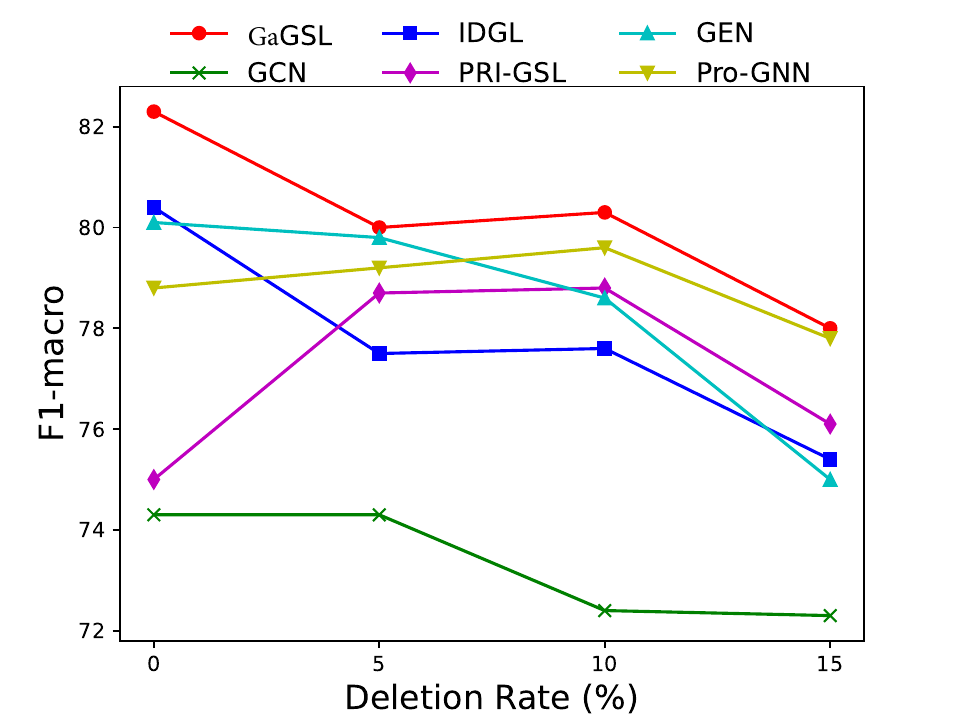}
    \centerline{\small{(b) Cora}}
    \label{fig:dele3}
    \end{minipage}
    \begin{minipage}[b]{0.3\textwidth}
    \centering
    \includegraphics[width=\textwidth]{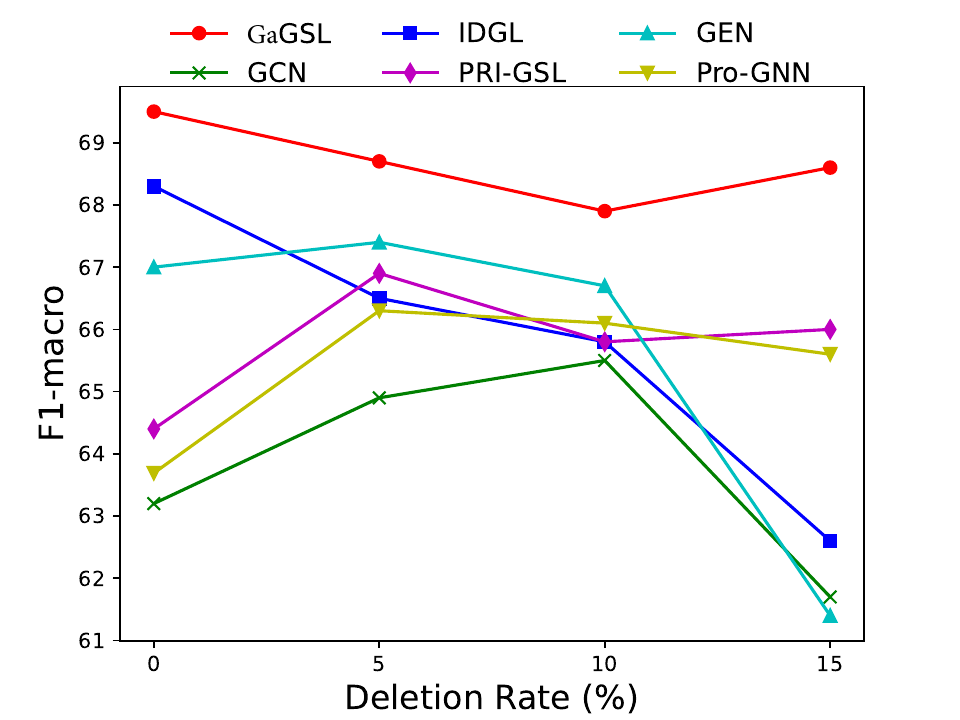}
    \centerline{\small{(c) Citeseer}}
    \label{fig:dele2}
    \end{minipage}
    \caption{Results of various models under random edge deletion on Cancer, Cora, and Citeseer datasets.}
    \label{fig3}
\end{figure*}

\begin{figure*}[htbp]
    \centering
    \begin{minipage}[b]{0.3\textwidth}
        \centering
        \includegraphics[width=\textwidth]{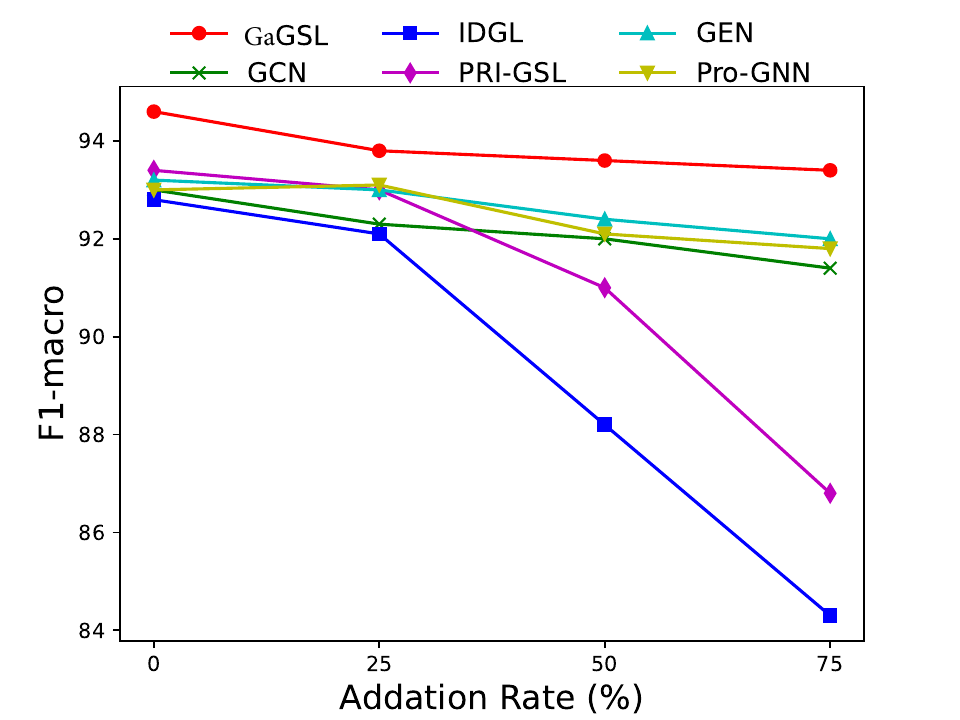}
        \centerline{\small{(a) Cancer}}
        \label{fig_add_cancer}
    \end{minipage}
    \begin{minipage}[b]{0.3\textwidth}
        \centering
        \includegraphics[width=\textwidth]{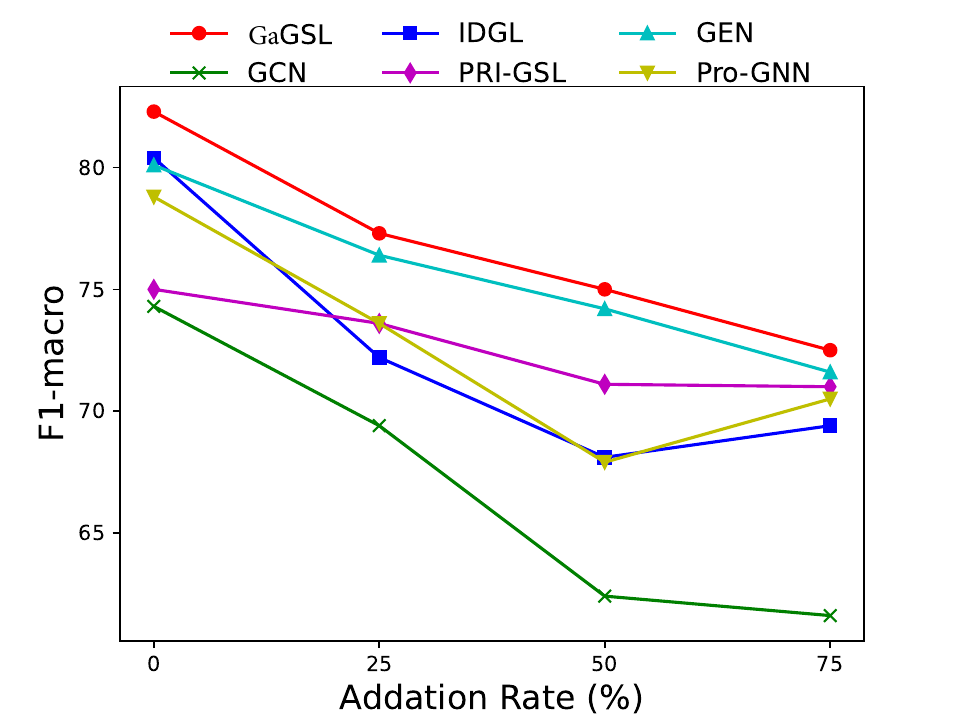}
        \centerline{\small{(b) Cora}}
        \label{fig_add_cora}
    \end{minipage}
    \begin{minipage}[b]{0.3\textwidth}
        \centering
        \includegraphics[width=\textwidth]{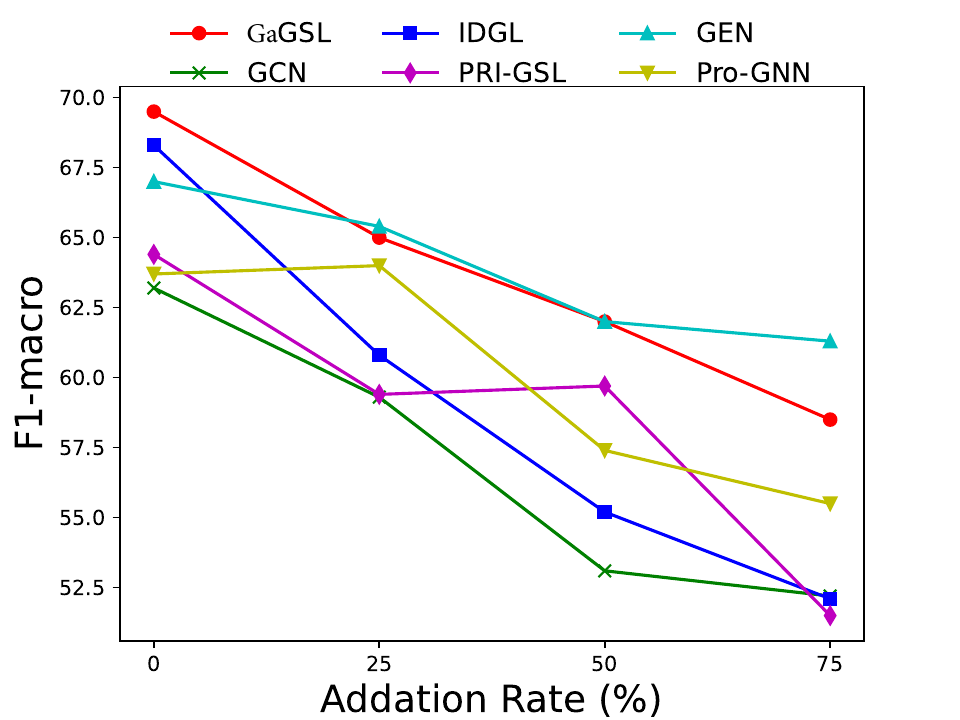}
        \centerline{\small{(c) Citeseer}}
        \label{fig_add_citeseer}
    \end{minipage}
    \caption{Results of various models under random edge addition on Cancer, Cora, and Citeseer datasets.}
    \label{fig4}
\end{figure*}

\begin{figure*}[htbp]
    \centering
    \begin{minipage}[b]{0.3\textwidth}
        \centering
        \includegraphics[width=\textwidth]{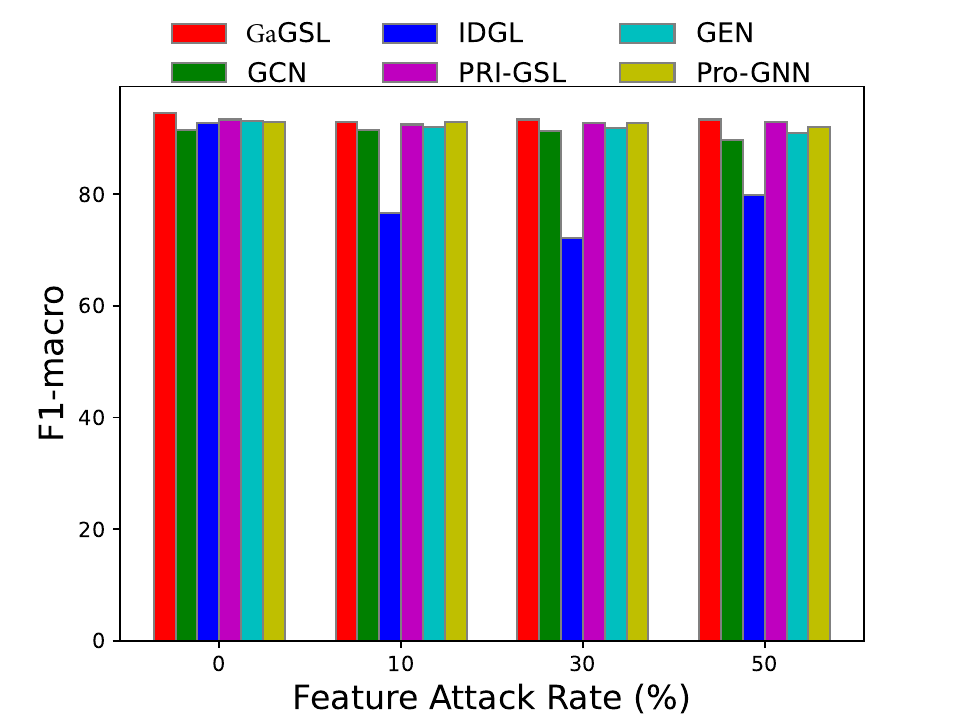}
        \centerline{\small{(a) Cancer}}
        \label{fig:add1}
    \end{minipage}
    \begin{minipage}[b]{0.3\textwidth}
        \centering
        \includegraphics[width=\textwidth]{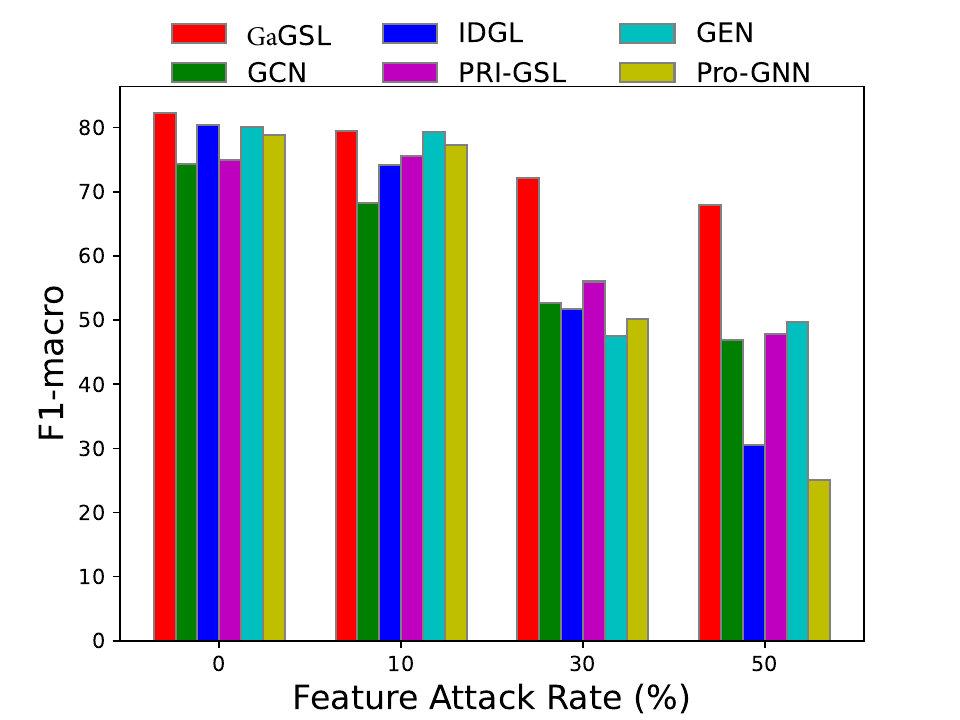}
        \centerline{\small{(b) Cora}}
        \label{fig:add3}
    \end{minipage}
    \begin{minipage}[b]{0.3\textwidth}
        \centering
        \includegraphics[width=\textwidth]{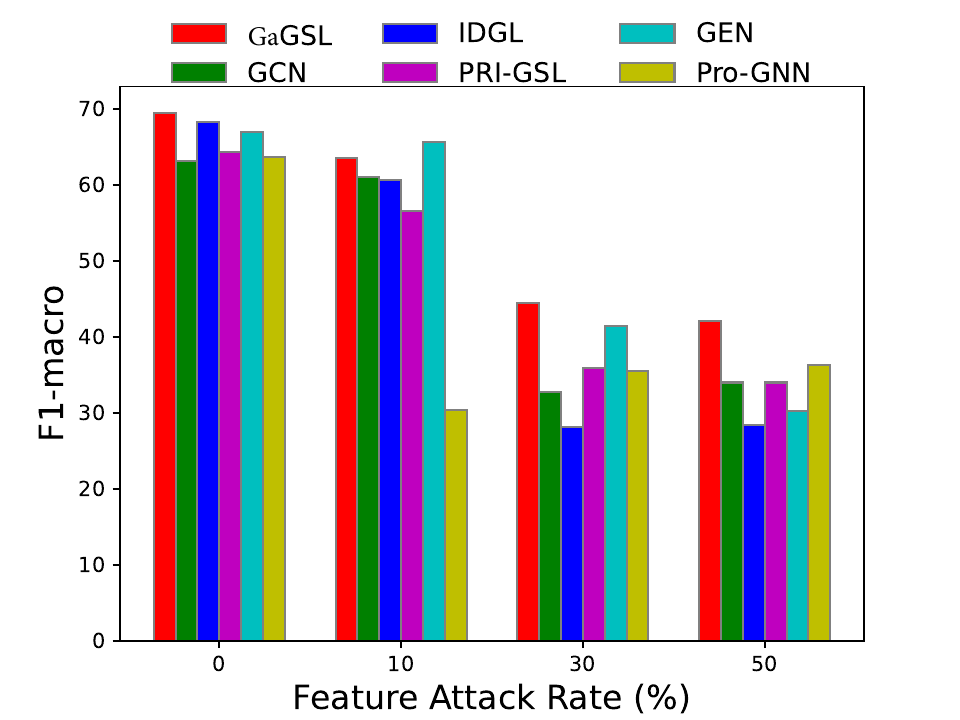}
        \centerline{\small{(c) Citeseer}}
        \label{fig:add2}
    \end{minipage}
    \caption{Results of various models under feature attack on Cancer, Cora, and Citeseer datasets.}
    \label{fig5}
\end{figure*}

\subsection{Defense Performance}
\noindent In this section, we perform a careful evaluation of various methods, specifically focusing on comparing GSL models. These models exhibit the ability to adapt the original graph structure, making them more robust in comparison to other GNNs. To ensure a comprehensive evaluation, we conduct attacks on both edges and features, respectively.

\subsubsection{Attacks on edges}
To attack edges, we generate synthetics dataset by deleting or adding edges on Cancer, Cora, and Citeseer following \cite{chen2020iterative}. Specifically, for each graph in the dataset, we randomly remove 5\%, 10\%, 15\% edges or randomly inject 25\%, 50\%, 75\% edges.
We select the poisoning attack \cite{zugner2018adversarial} and first generate the attacked graphs and subsequently train the model using these graphs. The experimental results are displayed in Figs. \ref{fig3} and \ref{fig4}.

As can be seen in Figs. \ref{fig3} and \ref{fig4}, GaGSL consistently outperforms all other baselines as the perturbation rate increases. On the Cancer and Citeseer datasets, the performance of GaGSL fluctuates lightly under different perturbation rates. 
On the Cora dataset, although GaGSL's performance declines with increasing perturbation rates, it still outperforms other baselines. Specifically, our model improves over vanilla GCN by 12.6\% and over other GSL methods by 0.8\%-11\% when 50\% edges are added randomly on Cora. This suggests that our method is more effective against robust attacks. We also find that the GSL method (i.e. Pro-GNN \cite{jin2020graph}, IDGL \cite{chen2020iterative}, GEN \cite{wang2021graph}, PRI-GSL \cite{sun2023self}) outperforms vanilla GCN at different perturbation rates on both the Cancer and Citeseer datasets, but on the Cancer dataset, IDGL is not as robust to attacks as vanilla GCN. The possible reason for this is the presence of labeling imbalances or outliers in the data, which may interfere with the model's ability to learn the correct graph structure and lead to performance degradation.

\subsubsection{Attacks on feature}
To attack feature, we introduce independent Gaussian noise to features following \cite{wu2020graph}. Specifically, we employ the mean of the maximum value of each node’s feature as the reference amplitude \(r\), and for every feature dimension of each node we introduce Gaussian noise \(\lambda\cdot r\cdot \epsilon\), where \(\lambda\) is the feature noise ratio, and \(\epsilon\sim N(0,1)\). We evaluate the models’ performance with \(\lambda \in \{0.1, 0.3, 0.5\}\). 
Additionally, we perform poisoning experiments and present the results in Fig. \ref{fig5}.

Similarly, we can observe from Fig. \ref{fig5} that in most cases, GaGSL consistently surpasses all other baselines and successfully resists feature attacks. As the feature attack rate increases, the performance of most methods decreases, but the decline rate of GaGSL is the slowest. For example, on the Cora dataset, as the perturbation rate increases from 0 to 50\%, the performance of GaGSL decreases by 13.3\%, while the performance of GCN decreases by 27.4\%.
In addition, we observed that under various feature attack scenarios, GaGSL consistently maintained top performance, whereas the rankings of other methods fluctuated. For example, on the Cora dataset, IDGL ranked second when the perturbation rate was 0\%, but dropped to fifth, fourth, and fifth at perturbation rates of 10\%, 30\%, and 50\%, respectively. Even more concerning, IDGL's performance was worse than that of GCN at perturbation rates of 30\% and 50\%.
This further demonstrates the strong robustness of GaGSL against feature attacks.
Combining the observations from Figs. \ref{fig3} and \ref{fig4}, we can conclude that GaGSL is able to approach the minimal sufficient structure, and thus demonstrates the capacity to withstand attacks on both graph edges and node features.

\begin{figure*}[t]

\begin{minipage}[b]{0.3\textwidth}
  \centering
  \centerline{\includegraphics[width=\textwidth]{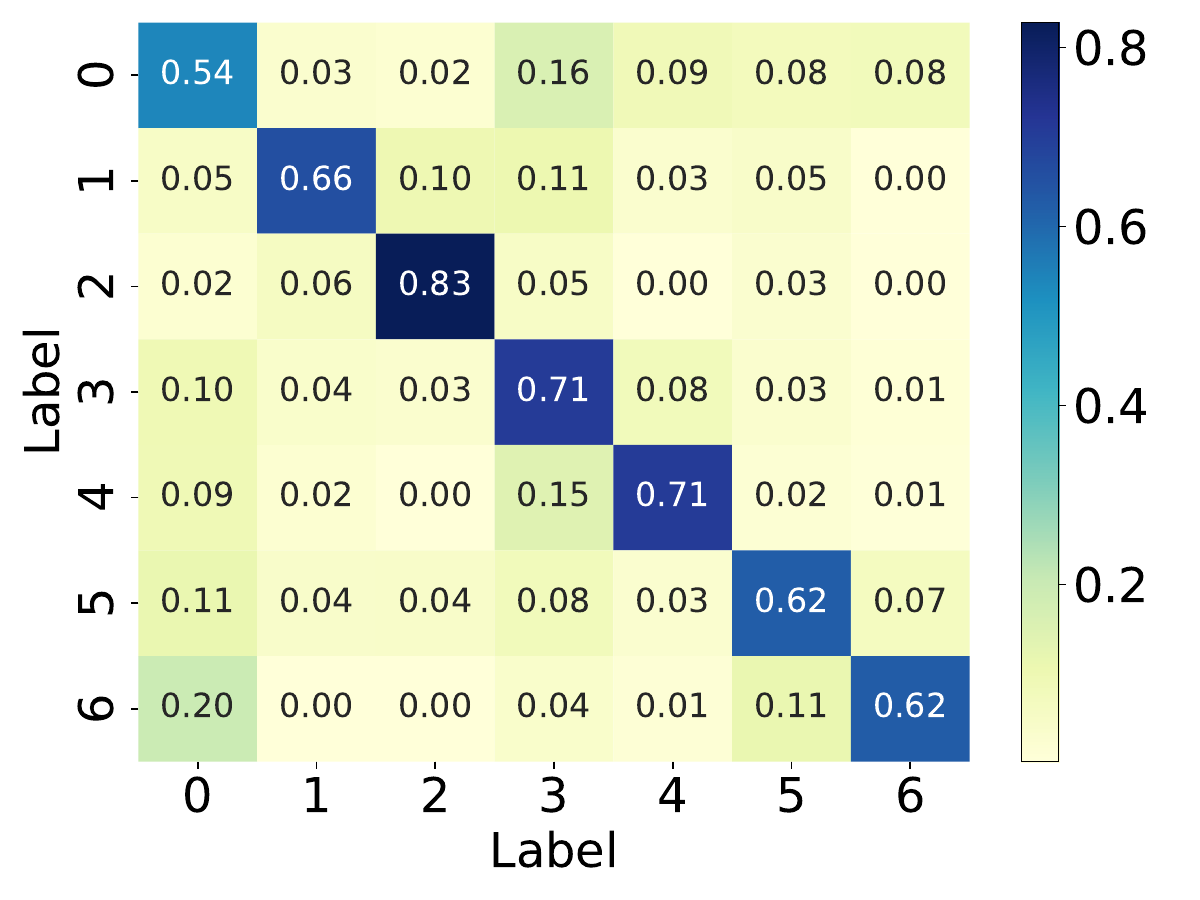}}
  \centerline{\small{(a) Original graph}}
\end{minipage}
\hfill
\begin{minipage}[b]{0.3\textwidth}
  \centering
  \centerline{\includegraphics[width=\textwidth]{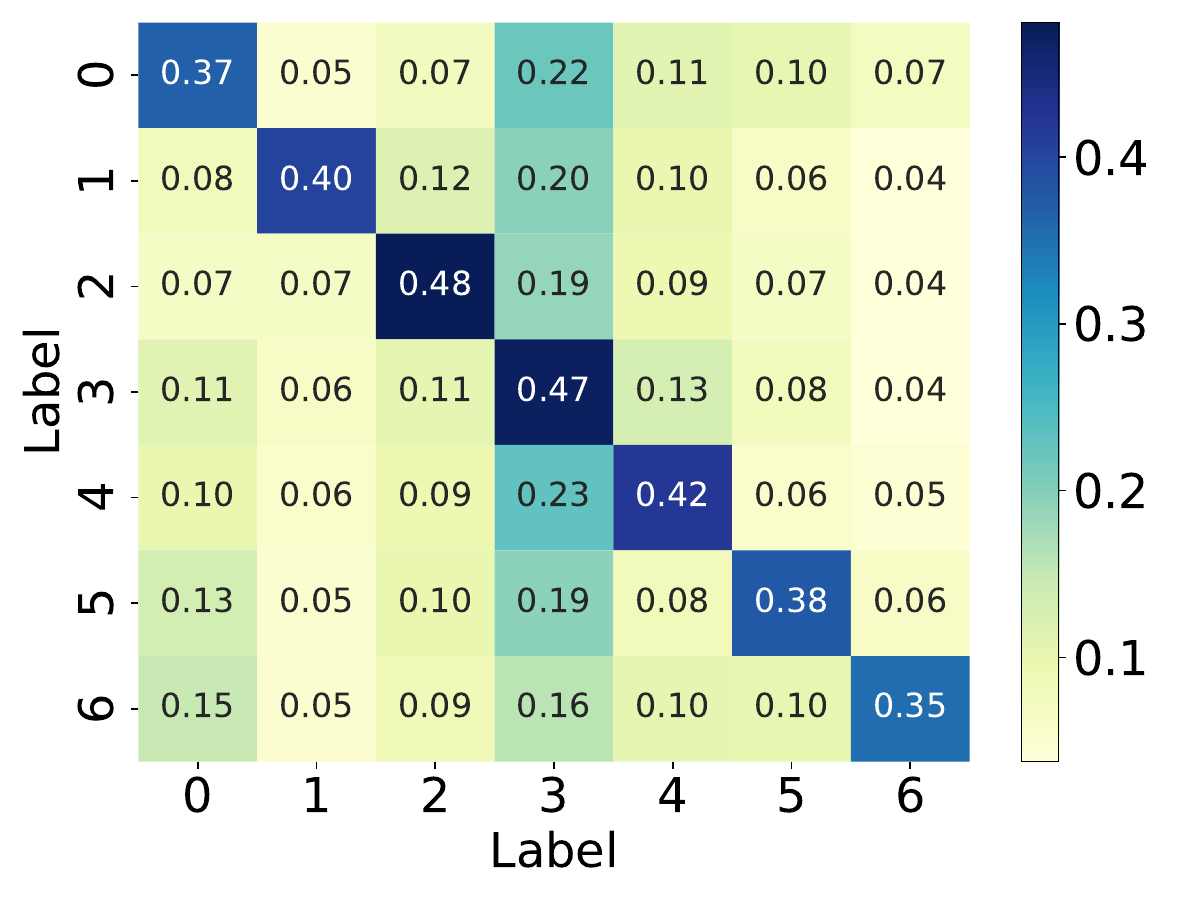}}
  \centerline{\small{(b) Perturbed graph}}
\end{minipage}
\hfill
\begin{minipage}[b]{0.3\textwidth}
  \centering
  \centerline{\includegraphics[width=\textwidth]{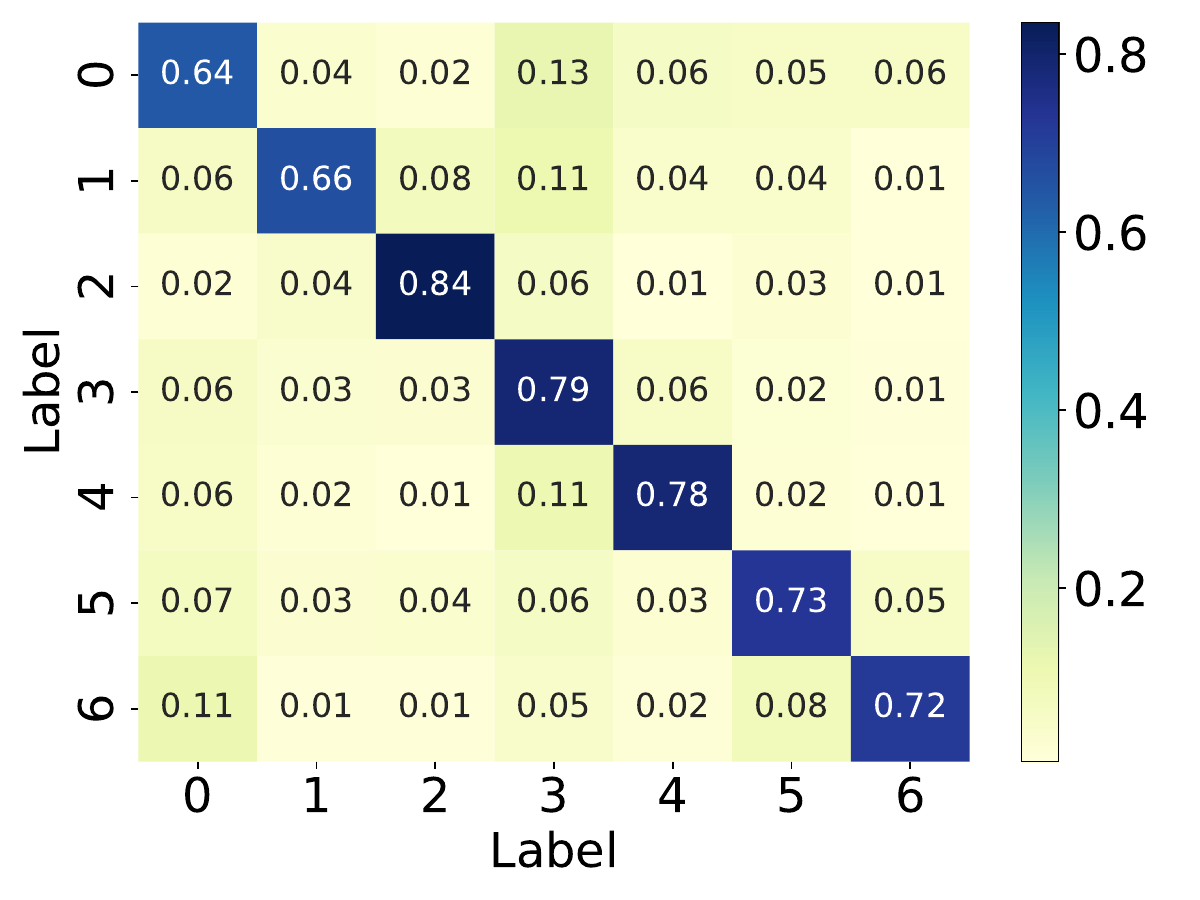}}
  \centerline{\small{(c) Learned graph}}
  \label{fig9(c)}
\end{minipage}
\caption{Heat maps for the probability matrices of the (a)
original graph, (b) Perturbed graph, and (c) Learned graph on Cora. Note that the color
shades of three maps represent different scales, as shown by
bars on the right.}
\label{fig9}
\end{figure*}

\begin{figure}[ht]
\begin{minipage}[b]{.48\linewidth}
  \centering
  \centerline{\includegraphics[width=\linewidth]{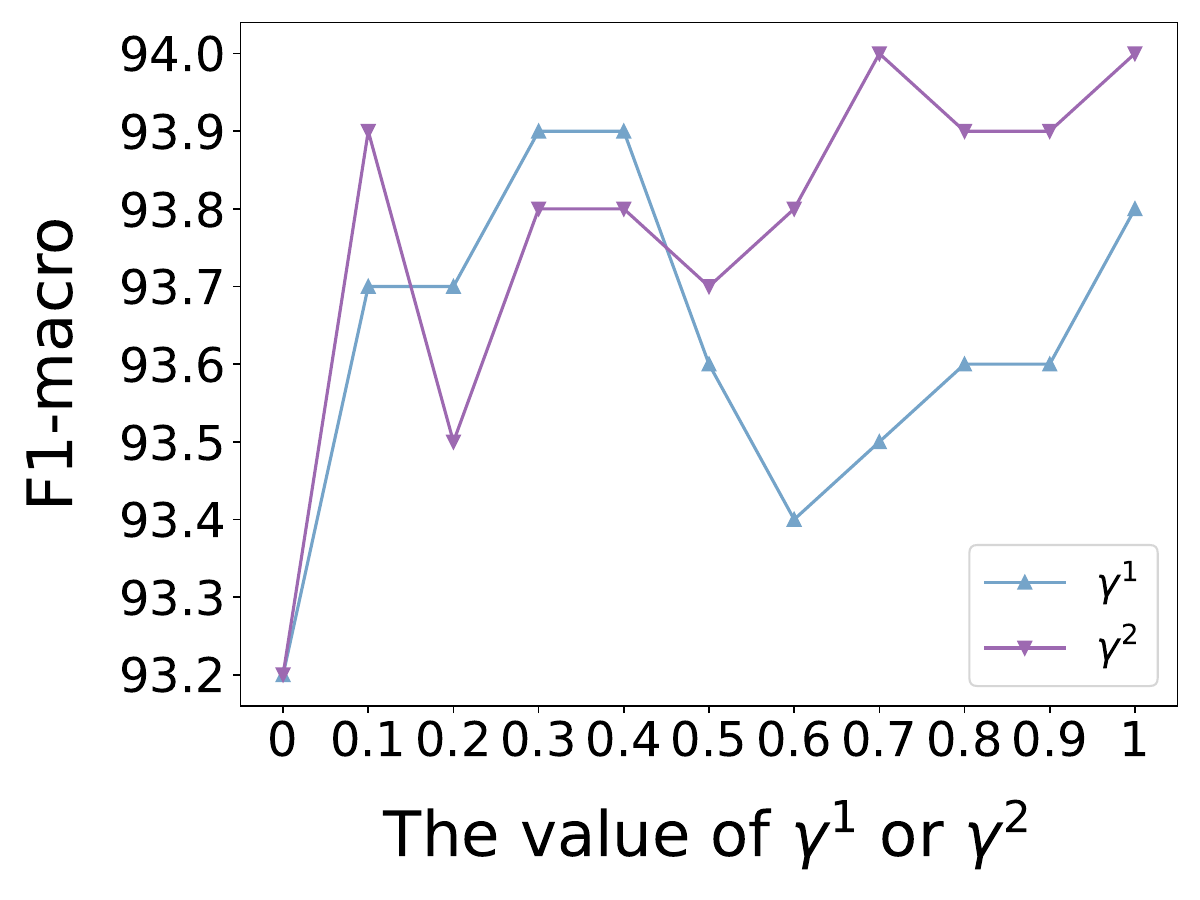}}
  \centerline{\small{(a) Polblogs}}
\end{minipage}
\hfill
\begin{minipage}[b]{.48\linewidth}
  \centering
  \centerline{\includegraphics[width=\linewidth]{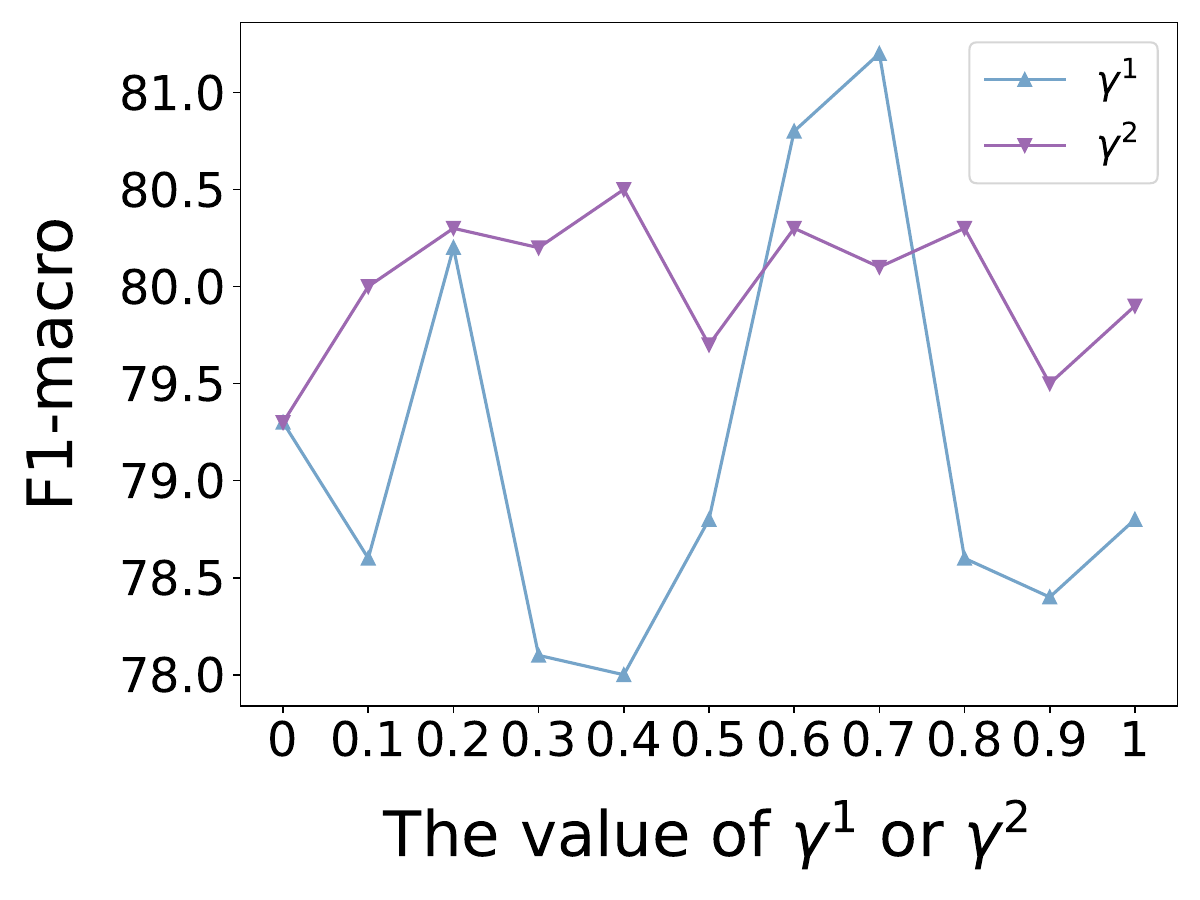}}
  \centerline{\small{(b) Cora}}
\end{minipage}
\caption{Impact of hyper-parameters \(\gamma^1\) and \(\gamma^2\) on Polblogs and Cora.}
\label{fig6}
\end{figure}
\begin{figure}[ht]
\begin{minipage}[b]{0.48\linewidth}
  \centering
  \centerline{\includegraphics[width=\linewidth]{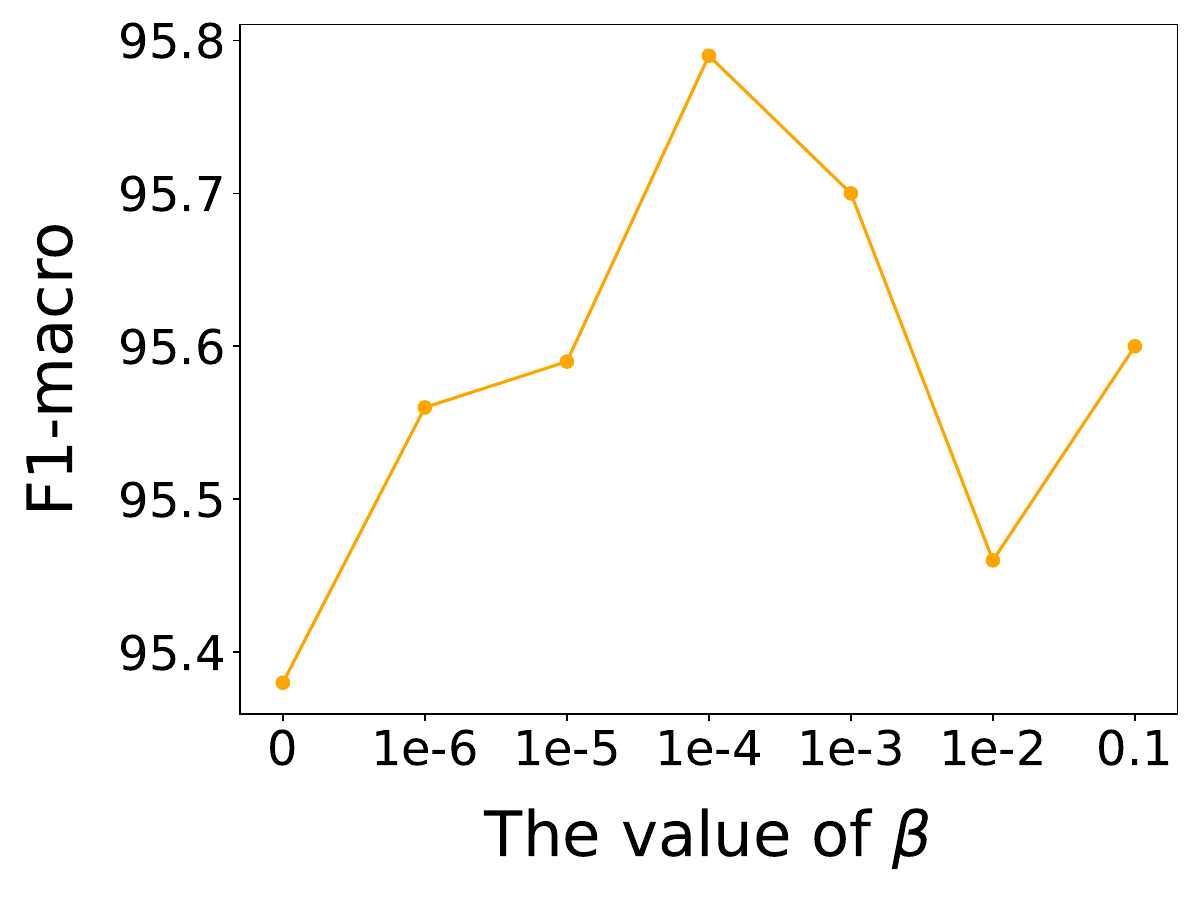}}
  \centerline{\small{(a) Polblogs}}
\end{minipage}
\hfill
\begin{minipage}[b]{0.48\linewidth}
  \centering
  \centerline{\includegraphics[width=\linewidth]{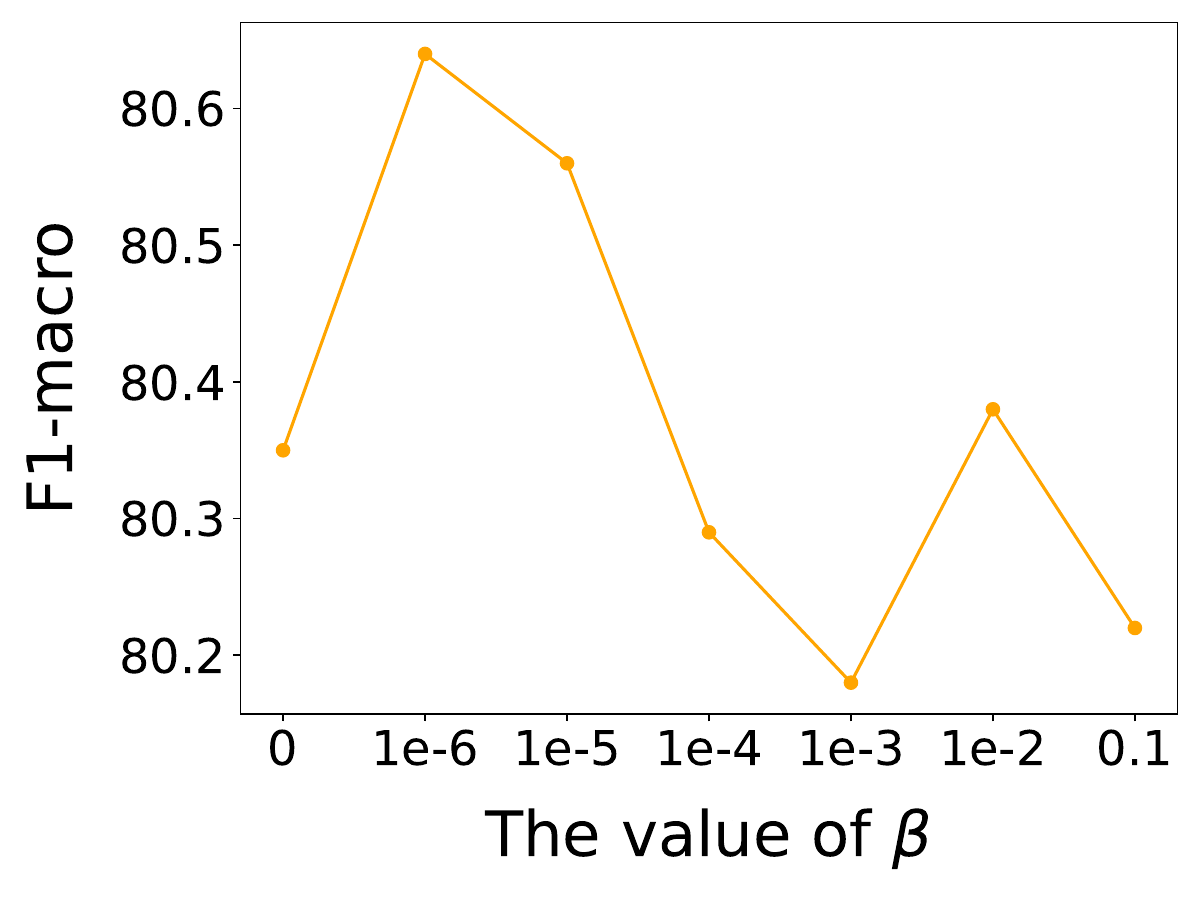}}
  \centerline{\small{(b) Cora}}
\end{minipage}
\caption{Impact of hyper-parameter \(\beta\) on Polblogs and Cora.}
\label{fig7}
\end{figure}

\subsection{Graph Structure Visualization} 
\noindent Here, we visualize the probability matrices of the original graph structure, the perturbed graph structure, and the graph structure learned by GaGSL and draw them in Fig. \ref{fig9}.

From the visualization, we can observe that in the perturbed graph structure, there exist noisy connections, as indicated by the higher probability of edges between different communities compared to the original graph structure.  These noisy connections degrade the quality of the graph structure, thereby reducing the performance of GNNs.  Additionally, we observed that, in contrast to the perturbed graph, the learned graph structure weakens the connections between communities and strengthens the connections within communities.

This observation is expected because GaGSL is optimized based on the GIB principle.  The GIB optimization allows GaGSL to effectively capture information in the graph structure that contributes to accurate node classification while constraining information that is label-irrelevant.  This optimization process ensures that the learned graph structure is robust against noisy connections and focuses on preserving the most informative aspects of the graph for the task at hand.

\subsection{Hyper-parameter Sensitivity}
\noindent In this subsection, we investigate the sensitivity of key hyper-parameters: combination coefficient \(\gamma^1\) in Eq. (\ref{eq12}), \(\gamma^2\) in Eq. (\ref{eq13}), as well as the balance parameter \(\beta\) in Eq. (\ref{eq31}). 
More concretely, we vary the value of \(\gamma^1\), \(\gamma^2\) and \(\beta\) to analyze their impact on the performance of our proposed model. 
We vary \(\gamma^1\) or \(\gamma^2\) or from 0 to 1, and  \(\beta\) from 0 to 0.1. For clarity, we report the node classification results on Cora and Polblogs datasets, as similar trends are observed across other datasets. The results are presented in Figs. \ref{fig6} and \ref{fig7}.

As can be observed from Fig. \ref{fig6}, by tuning the value of \(\gamma^1\) (or \(\gamma^2\)), we can achieve better node classification performance compared to when \(\gamma^1\) (or \(\gamma^2\)) is set to 0.
This precisely demonstrates that our designed structure estimator is capable of capturing the complex correlations and semantic similarities between nodes.

Observing Fig. \ref{fig7}, it is evident that as the value of \(\beta\) increases, the performance of GaGSL first rises to a peak and then declines. When \(\beta\) is small, \(\mathcal{L}_{ML}\) has a minimal impact on the total loss function \(\mathcal{L}\), and \(\mathcal{L}_{cls}\) dominates the model's training. At this stage, increasing \(\beta\) enhances the influence of \(\mathcal{L}_{ML}\), thereby helping the model better constrain the label-irrelevant information from \(G_{r1}\) and \(G_{r2}\) in \(G\), thus improving performance. However, if \(\beta\) becomes too large, the model may excessively focus on \(\mathcal{L}_{ML}\) and neglect \(\mathcal{L}_{cls}\) during training, leading to a decline in performance. Overall, an appropriate value of \(\beta\) can effectively balance the influence of \(\mathcal{L}_{cls}\) and \(\mathcal{L}_{ML}\), thereby enhancing the model's performance.

\begin{figure}[tbp]
\begin{minipage}[b]{.49\linewidth}
  \centering
  \centerline{\includegraphics[width=\linewidth]{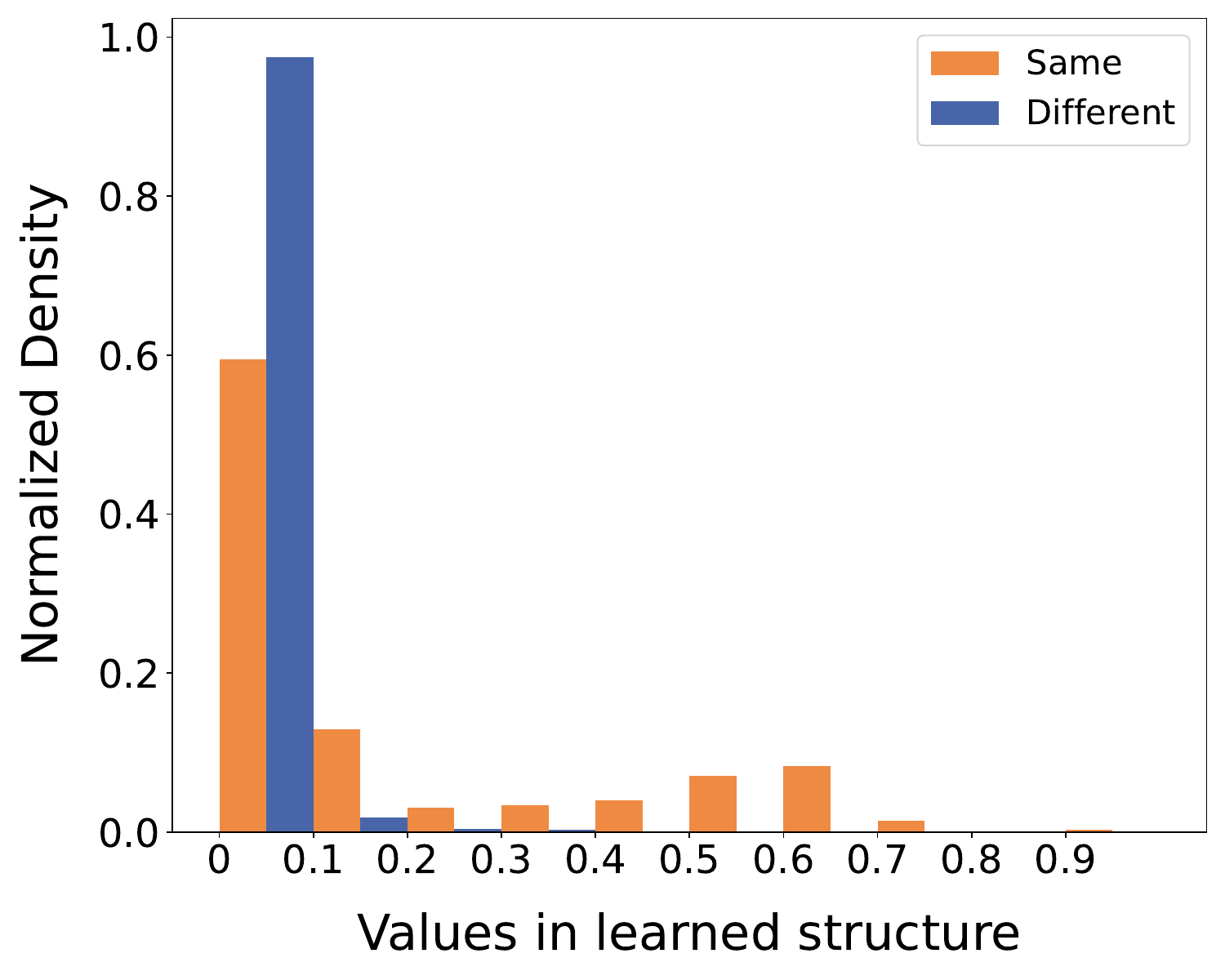}}
  \centerline{\small{(a) Citeseer}}
\end{minipage}
\hfill
\begin{minipage}[b]{0.49\linewidth}
  \centering
  \centerline{\includegraphics[width=\linewidth]{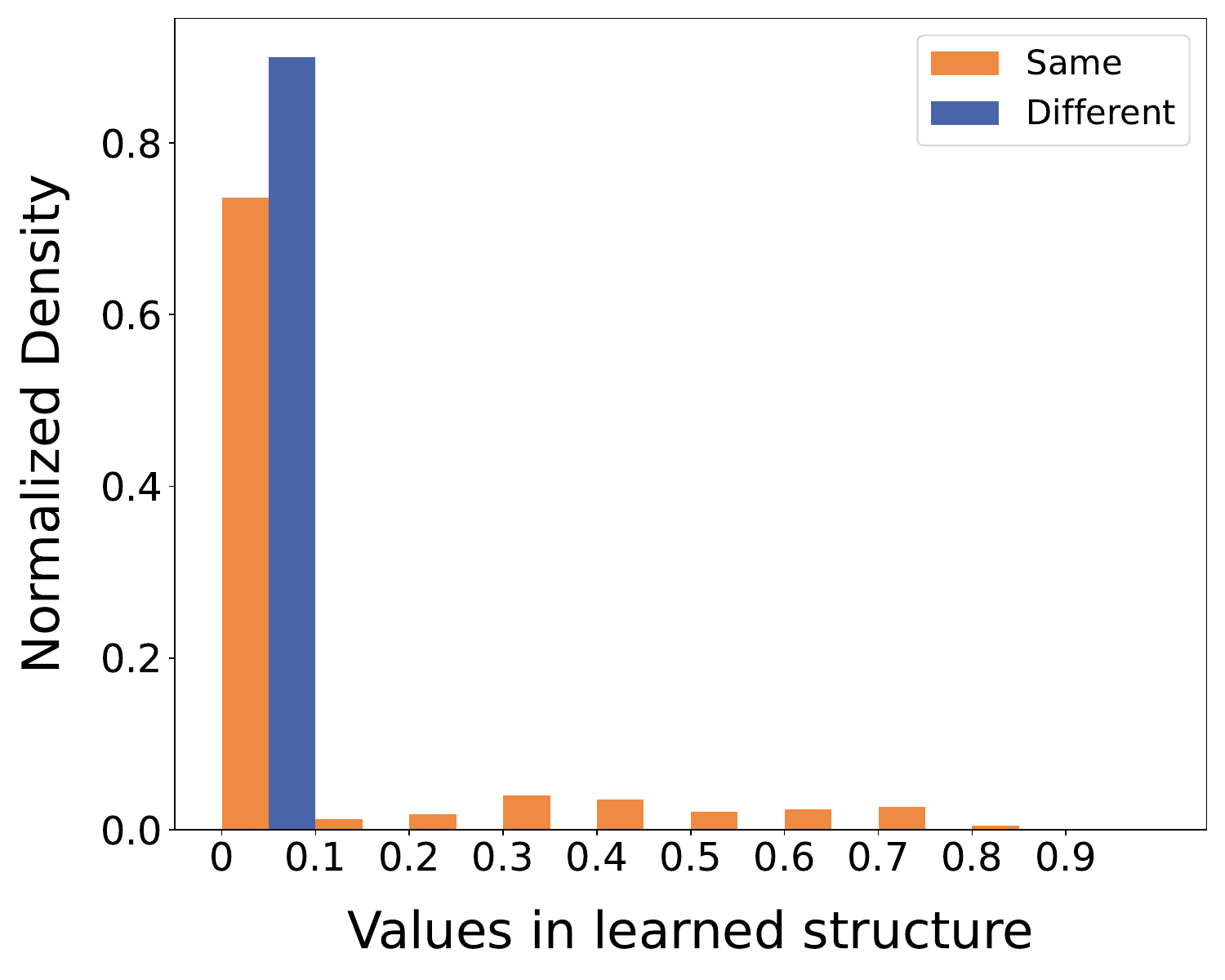}}
  \centerline{\small{(b) Polblogs}}
\end{minipage}
\caption{Standard histogram of learned adjacency matrix weights on Citeseer and Polblogs}
\label{fig8}
\end{figure}

\subsection{Values in Learned Structure} 
\noindent To further elucidate the distribution of edge weights between community nodes, we categorize the edges into two teams: edges connecting nodes within the same communities and edges connecting nodes across different communities. In Fig. \ref{fig8}, we showcase the normalized histograms of the learned structure values (weights) on Citeseer and Polblogs datasets.

From the histograms, it is apparent that the weights of edges between different communities are predominantly concentrated in the first bin (less than 0.1). In contrast, the weights of edges within the same communities are distributed not only in the first interval but also in higher bins. This distribution demonstrates that GaGSL effectively differentiates between inter-community and intra-community edges when assigning weights. Consequently, this ability to assign appropriate weights enhances the performance and robustness of graph representations.

\section{\label{Con}Conclusion}
\noindent In this paper, we proposed a novel GSL method GaGSL, which tackles the challenge of learning a graph structure for node classification tasks guided by the GIB principle.
The GaGSL method consists of three parts: global feature and structure augmentation, structure redefinition, and GIB guidance. We first mitigate the inherent bias of relying solely on a single original structure by global feature and structure augmentation.
Subsequently, we design the structure estimator with different parameters to refine and optimize the graph structure. Finally, The optimization of the final graph structure is guided by the GIB principle. Experimental results on various datasets validate the superior effectiveness and robustness of GaGSL in learning compact and informative graph structure.

Future work will delve into further optimizing the GaGSL model to address label noise and data imbalance.  Although this paper effectively mitigates structural noise, challenges posed by label noise and data imbalance remain unresolved.  Therefore, upcoming research will concentrate on devising robust strategies to manage label noise and handle data imbalance issues.  By tackling these challenges, we aim to ensure that the model achieves enhanced performance and robustness in real-world scenarios.

\bibliographystyle{IEEEtran}
\bibliography{ref.bib}

\vfill

\end{document}